%% file: acl2023.tex
\pdfoutput=1

\documentclass[11pt]{article}

\usepackage[]{ACL2023}

\usepackage{times}
\usepackage{amsmath}
\usepackage{latexsym}
\usepackage{graphicx}
\usepackage{booktabs}
\usepackage{rotating}
\usepackage{adjustbox}
\newcommand*\rot{\rotatebox{90}}

\usepackage{multirow}
\usepackage{colortbl}
\usepackage{paralist}
\usepackage{hhline} 
\usepackage{highlight}
\usepackage{pifont}
\usepackage{booktabs, multirow} 

\usepackage{textcomp}
\newcommand{\cmark}{\ding{51}}%
\newcommand{\xmark}{\color{lightgray}\ding{55}}%

\usepackage[T1]{fontenc}

\usepackage[utf8]{inputenc}
\usepackage{subfig}

\usepackage{microtype}

\usepackage{inconsolata}
\input{internal_comments}
\usepackage{xcolor}
\definecolor{newblue}{rgb}{0,0.1,.8}
\definecolor{neworange}{rgb}{1,0.5,.1}

\title{Can Incidental Bilingualism Help Explain the Translation Capabilities \\ of Large Language Models?}
\title{\textit{Searching for Needles in a Haystack:} \\
On the Role of Incidental Bilingualism in PaLM's Translation Capability
}

\author{Eleftheria Briakou \\
  \texttt{ebriakou@cs.umd.edu}\And
  Colin Cherry \\
  \texttt{colincherry@google.com} \And
  George Foster \\
  \texttt{fosterg@google.com}}

\begin{document}
\maketitle


\begin{abstract}

    Large, multilingual language models exhibit surprisingly good zero- or few-shot machine translation capabilities, despite having never seen the intentionally-included translation examples provided to typical neural translation systems. We investigate the role of \textit{incidental bilingualism}---the unintentional consumption of bilingual signals, including translation examples---in explaining the translation capabilities of large language models, taking the Pathways Language Model (PaLM) as a case study. We introduce a mixed-method approach to measure and understand incidental bilingualism at scale. We show that PaLM is exposed to over $30$ million translation pairs across at least $44$ languages. Furthermore, the amount of incidental bilingual content is highly correlated with the amount of monolingual in-language content for non-English languages.
    We relate incidental bilingual content to zero-shot prompts and show that it can be used to mine new prompts to improve PaLM's out-of-English zero-shot translation quality. Finally, in a series of small-scale ablations, we show that its presence has a substantial impact on 
    translation capabilities, although this impact diminishes with model scale.
\end{abstract}

\input{S1.Introduction}

\input{S2.Background}

\input{S3.Measuring}

\input{S4.Analyzing}

\section{Conclusion}

We explore the role of incidental bilingualism---the unintentional consumption of bilingual signals---in PaLM's translation capabilities. We introduce a mixed-method approach that alternates between quantitative and qualitative analyses to measure and understand incidental bilingualism at scale by processing $780$ billion tokens. Our work shows that PaLM consumes a significant amount of bilingual text: $1.4\%$ of training instances in natural language are bilingual. At the same time, it is naturally exposed to translation signals, having seen more than $30$ million translation pairs in $44$ languages paired with English. Furthermore, we extrinsically evaluate the quality of these translations, showing that they can be used to train supervised models that roughly match the quality of equal amounts of WMT data. Finally, we show that incidental bilingualism connects to the machine translation capabilities of PaLM. First, we show that data-driven prompts extracted from incidental translations can improve the zero-shot abilities of PaLM when translating out of English by $14$ chrF on average. Second, we provide empirical evidence that bilingual and translation signals can partially explain the translation capabilities of smaller-scale \textsc{llm}s.

\section*{Limitations}
Our findings should be interpreted considering a series of problem definitions and design choices. 
First, our quantitative results on measuring incidental bilingualism at scale
are subject to language identification, sentence splitting, and mining errors. 
Our qualitative analysis for the English-French language pair revealed that those errors are reasonably small (see \S\ref{sec:qual2}). However, we expect the accuracy of our tools to vary across languages and, crucially, exhibit unanticipated failure modes on web text and low-resource languages~\cite{caswell-etal-2020-language}. 
Second, our findings are restricted to quantifying bilingualism and translations within a limited set of language pairs and only paired with English. Thus, by problem definition, we are limited to computing a lower-bound estimate on incidental bilingualism of PaLM.
The above limitations should also be taken into consideration when interpreting our ablation results. Although we attempted to remove most bilingual signals in our series of \textsc{mt} experiments, it is still possible that bilingualism slips through due
to either model errors or due to bilingual signals beyond our focus set of languages.
Finally, any results and findings of our work are restricted to PaLM; the single \textsc{llm} studied in this work. However, our finer-grained analysis (see Table~\ref{tab:per_source_counts} of Appendix~\ref{sec:appendix})
reveals that \textit{incidental bilingualism}, including translation signals, is observed across various data sources (e.g., webpages, books, etc.) that are commonly included in the training data of other popular \textsc{llm}s.


\section*{Acknowledgements}
We thank 
Jiaming Luo,
Julia Kreutzer,
Orhan Firat,
Xavier Garcia,
Markus Freitag,
Sweta Agrawal, 
Marine Carpuat,
Elijah Rippeth,
and the anonymous reviewers for their helpful and constructive comments.


\bibliography{anthology,custom}
\bibliographystyle{acl_natbib}
\newpage
\appendix
\input{S5.Appendix}

\end{document}

%% file: internal_comments.tex
\usepackage{xcolor}

\newcommand{\ignore}[1]{}

%% file: S1.Introduction.tex
\section{Introduction}

Recent work has shown that large language models (\textsc{llm}s) exhibit impressive capabilities in performing various natural language generation tasks, even in the zero-shot paradigm. In particular, such models have shown interesting machine translation (\textsc{mt})  capabilities~\cite{NEURIPS2020_1457c0d6, Chowdhery2022PaLMSL,vilar2022prompting}---especially when translating into English, despite never having been \textit{explicitly} and \textit{intentionally} exposed to translation data in the way their supervised counterparts are.
This raises the question: where do these translation capabilities come from?

We hypothesize that the translation capabilities of \textsc{llm}s connect to \textit{incidental bilingualism}: the unintentional consumption of bilingual text within a single training instance. To test this hypothesis, we take PaLM~\cite{Chowdhery2022PaLMSL}---a $540$-billion parameter Transformer language model---as a case study. We first conduct a large-scale analysis of its training data in order to characterize the nature and quantity of bilingual text, then perform experiments to assess the impact of this text on translation performance.

To measure incidental bilingualism at scale, we develop a processing pipeline that
alternates between quantitative and qualitative analysis (\S\ref{sec:measuring}): first detect bilingual versus monolingual text using a language tagger, then qualitatively analyze the nature of bilingual text, and finally measure the amount of translation data within bilingual instances.
Our analysis spans $44$ languages, for which we study bilingualism paired with English.
Our findings are:
\begin{itemize}
    \item In all, $1.4\%$ of \textsc{palm}'s training instances are detected as bilingual, while $0.34$\% 
    contain at least one translated sentence pair. We were able to mine such pairs across all languages studied; therefore,  none of these languages is truly zero-shot in the context of translation. 
    \item The number of monolingual instances in a language is predictive of the number of instances containing bilingual or translation content for that language (paired with English).
\end{itemize}

After establishing that both bilingual and translation content are incidentally consumed during PaLM's training, we study how they connect to its \textsc{mt} capabilities (\S\ref{sec:analyzing}). We run a series of training and prompting experiments and found that:
\begin{itemize}
    \item Prompting the full PaLM model with alternative, data-driven prompts improves out-of-English zero-shot translation by $14$ chrF points on average across languages, indicating that its zero-shot translation capabilities were underestimated due to sub-optimal prompts.
    \item Ablating detected translation pairs with smaller versions of PaLM
    has a dramatic effect on the translation capabilities of 1B-parameter models for high-resource languages, reducing average into-English zero-shot results by $7.4$ \textsc{bleu} and $5$-shot results by $5.9$ \textsc{bleu}. The effect falls off but remains notable ($+2$-$3$ \textsc{bleu} across several conditions) as we scale to $8$B-parameter models.
\end{itemize}

%% file: S2.Background.tex
\section{Related Work}

\paragraph{Translation Capabilities of LLMs} Large-scale generative language models, such as \textsc{gpt-3}~\cite{NEURIPS2020_1457c0d6}, PaLM~\cite{Chowdhery2022PaLMSL}, and \textsc{xglm}~\cite{Lin2021FewshotLW} have been shown to exhibit translation capabilities, despite not being explicitly trained to translate. These capabilities are surprisingly strong, particularly when translating into English with few-shot examples.
One explanation for this behavior is that it results from incidental multitask learning \cite{Radford2019LanguageMA,Sanh2021MultitaskPT}. This hypothesis has not been explored for \textsc{mt}, where recent work has mostly focused on 
improving \textsc{llm} translation capabilities by optimizing few-shot prompting strategies~\cite{vilar2022prompting, Agrawal2022IncontextES}.
Rather than trying to improve translation quality for \textsc{llm}s, our goal is to understand where their translation abilities stem from by tracing them back to the properties of the pretraining data.

\paragraph{Large-Scale Data Analysis} \textsc{llm}s rely on massive amounts of unlabeled corpora for training. These corpora are primarily acquired by combining heterogeneous online resources (e.g., Wikipedia, Web forums, Common Crawl, etc.)---whose properties are usually unknown. Recent work on large-scale analysis has shed some light: 
\citet{dodge-etal-2021-documenting} analyze C4~\cite{Raffel2019ExploringTL}---a dataset created from a snapshot of Common Crawl---and show that it contains machine generated texts as well as evaluation samples from commonly used \textsc{nlp} benchmarks;  
\citet{kreutzer-etal-2022-quality} manually audit the quality of multilingual datasets and find systematic quality issues amongst popular pretraining datasets.
Most related to our work, \citet{Blevins2022LanguageCH} show that popular corpora routinely used for training English-only \textsc{llm}s contain a non-negligible amount of non-English text, which helps explain their cross-lingual capabilities. Their manual analysis of corpus subsamples covers several bilingual categories, including a translation category. But where analysis of bilingualism is a side result of their work, it is our primary contribution. We extend their work by proposing automatic tools to quantify bilingualism at scale and directly relate it to \textsc{llm} translation performance.

\paragraph{Eliciting Knowledge from LLMs} Prompting language models to elicit knowledge acquired during pre-training has received a lot of research interest. \citet{petroni-etal-2019-language} show that \textsc{llm}s  can recall factual knowledge by answering queries structured as cloze statements. 
\citet{jiang-etal-2020-know} further show that query-based prompts outperform manually created cloze statements, suggesting that the latter provide a lower bound estimate on the actual abilities of \textsc{llm}s. Follow-up work confirms those findings by suggesting better prompts with automatic generation methods~\cite{shin-etal-2020-autoprompt} or prompt engineering~\cite{Reynolds2021PromptPF}. We similarly explore how to extract translation knowledge from \textsc{llm}s using data-driven prompts. 

%% file: S3.Measuring.tex
\section{Measuring \& Understanding Incidental Bilingualism}\label{sec:measuring}

\begin{figure}[!t]
    \centering
    \includegraphics[scale=0.15]{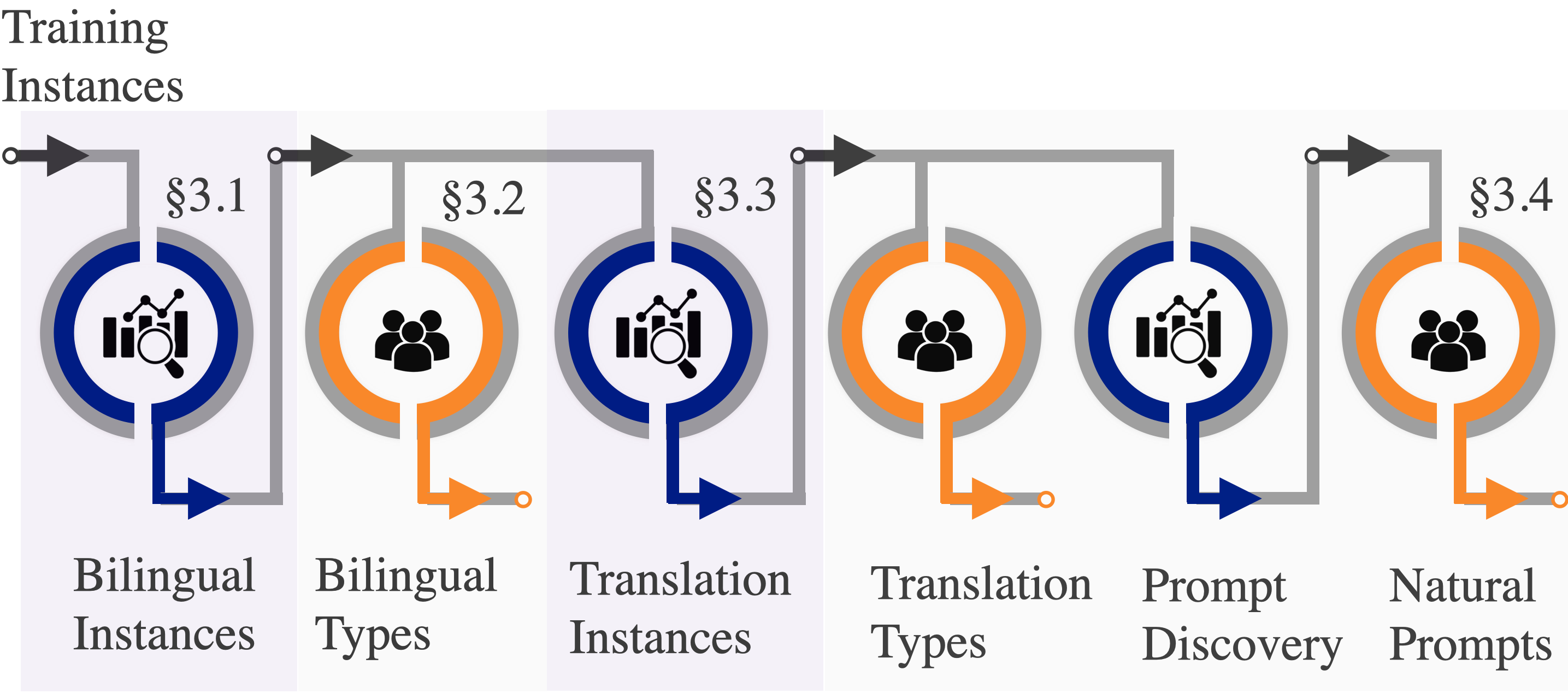}
    \caption{A mixed-method approach to measure and understand incidental bilingualism at scale. We alternate between \textcolor{newblue}{quantitative} and \textcolor{neworange}{qualitative} steps to detect (\S3.1) and analyze (\S3.2) bilingual instances, then detect (\S3.3) and analyze (\S3.4) translation instances. }
    \label{fig:mixed}
\end{figure}

We introduce a mixed-method approach~\cite{creswell, shorten} to measure and understand \textit{incidental bilingualism}---the unintentional consumption of bilingual signals---at scale. Since we expect bilingual signals to be rare, we explore the huge data space by alternating between quantitative and qualitative steps, with results from each step complementing and informing one another (Figure~\ref{fig:mixed}).  The quantitative steps play the role of inducing a smaller-scale focus space to study, while the qualitative steps 
provide insights into the nature of bilingual signals.

\paragraph{Preliminaries}
PaLM's pretraining dataset consists of $780$ billion tokens from a mixture of multilingual sources (social media conversations ($50\%$), filtered webpages ($27\%$), and Wikipedia ($4\%$)), presumably English sources (books ($13\%$) and news articles ($1\%$)), and source code ($5\%$).
PaLM was trained on $2{,}048$-subword-token examples formed by concatenating and truncating documents.
As PaLM is a multi-source \textsc{lm}, a document may be a web page, a book, or a conversation, depending on the source.
%
Our primary units for data analysis are \textit{instances} we created by splitting training examples along document boundaries. As such, each instance is either a complete document or a contiguous fragment of one, up to $2{,}048$ tokens in length. A more detailed discussion of instances is given in Appendix~\ref{sec:appendix_analysis}.

We study bilingualism between English and $44$ other languages. We choose language pairs that: a) are supported by our language identification models, and b) have \textsc{flores}-$101$~\cite{goyal-etal-2022-flores} evaluation data. We divide languages into high, medium, and low-resource groups according to their monolingual instance counts, as shown below:

\begin{table}[!ht]
    \centering
    \scalebox{0.8}{
    \begin{tabular}{lp{0.85\linewidth}}
    \textbf{\textsc{high}} & \textsc{fr},  \textsc{de},  \textsc{es},  \textsc{it} \\
    \textbf{\textsc{medium}} &	 \textsc{pt},  \textsc{ru},  \textsc{zh},  \textsc{ja},  \textsc{ar},  \textsc{id},  \textsc{ko},  \textsc{vi},  \textsc{fa},  \textsc{sr},  \textsc{uk}\\
    \textbf{\textsc{low}} &  \textsc{ps},  \textsc{hy},  \textsc{iw},  \textsc{bg},  \textsc{kk},  \textsc{be},  \textsc{hi},  \textsc{ur},  \textsc{el},  \textsc{th},  \textsc{mk},  \textsc{ky},  \textsc{bn},  \textsc{ka},  \textsc{tg},  \textsc{sd},  \textsc{ne},  \textsc{ta}, \textsc{mn},  \textsc{pa},  \textsc{te},  \textsc{ml},  \textsc{mr},  \textsc{am},  \textsc{my},  \textsc{kn},  \textsc{km},  \textsc{gu},  \textsc{lo}\\
    \end{tabular}}\vspace{-1.em}
\end{table}


\subsection{Detecting Bilingual Instances}\label{sec:quant1}

Our first goal is to automatically detect all training instances that contain bilingual text without presupposing a specific granularity for bilingualism. To that end, we use \textsc{cmx}~\cite{zhang-etal-2018-fast-compact}---a language identification model for codemixed texts---to produce a sequence of token-level language tags for each training instance. An instance is labeled as bilingual if it contains at least two contiguous segments in different languages, each consisting of at least $N$ consecutive identical language tags.
Instances with more than two languages are interpreted as bilingual, as discussed in Appendix \ref{sec:appendix_cmx}. 
One of the two languages must always be English, both to simplify our analysis and to work within the limits of the \textsc{cmx} tool.

\begin{figure}[!t]
    \centering
    \includegraphics[scale=0.2]{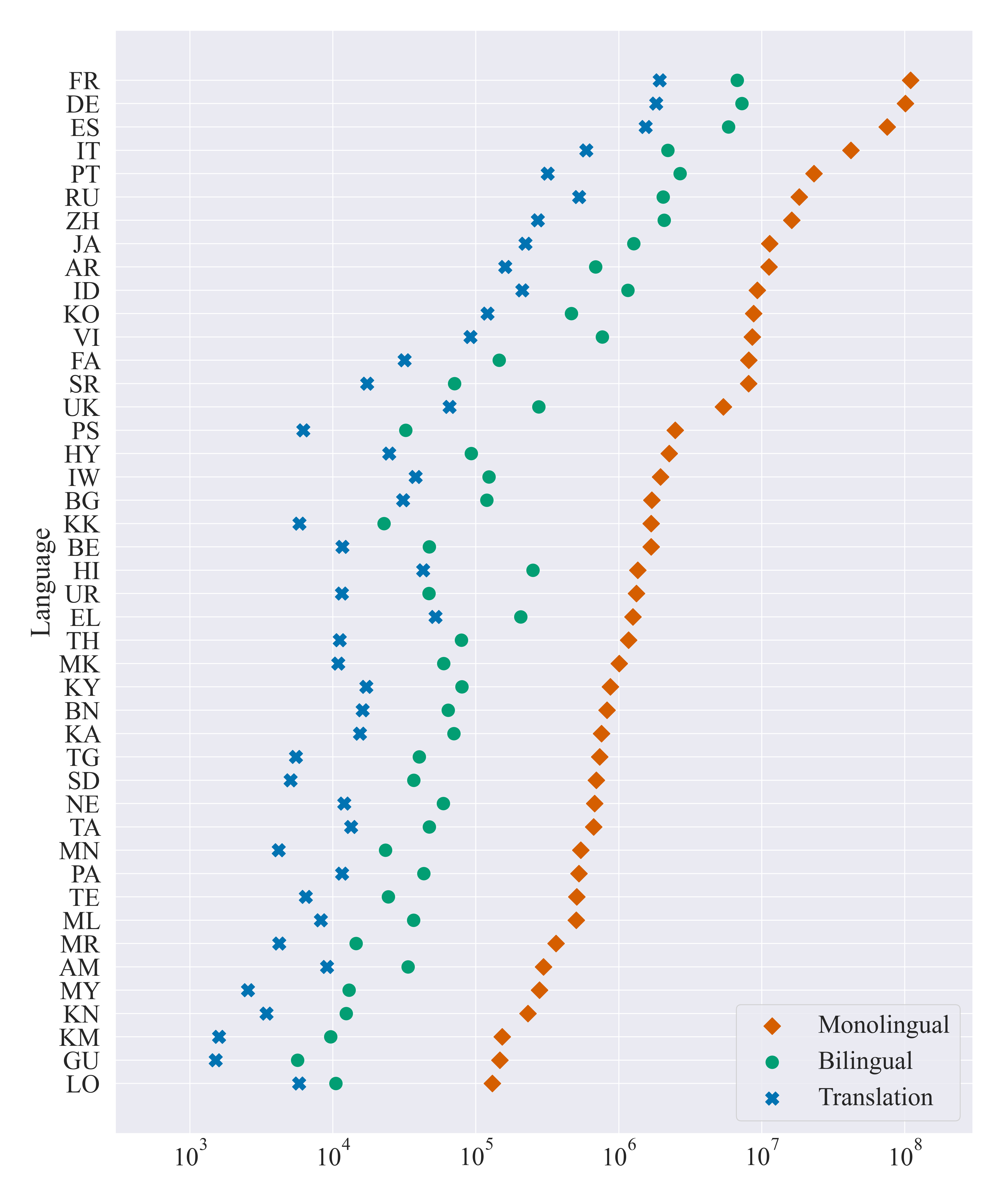}
    \caption{Number of monolingual, bilingual, and translation instances detected within PaLM's training data. PaLM consumes bilingual signals, including translation examples, across (at least) $44$ languages.}
    \label{fig:bilingual_stats}
\end{figure}

\paragraph{Findings} Figure~\ref{fig:bilingual_stats} presents the per-language monolingual and bilingual instance counts. We include raw counts per language in Table~\ref{tab:full_stats}.
%
We observe that across the languages studied, PaLM consumes bilingual instances that, in total, account for $1.4\%$ of its training instances.

\subsection{Characterizing Bilingual Instances}\label{sec:qual2}

Next, we turn to understanding the nature of bilingual instances detected by the above procedure. 
To make manual analysis easier, we used the \texttt{KnowYourData} tool\footnote{\url{https://knowyourdata.withgoogle.com}} to highlight spans of the less frequent language in each bilingual instance.

\begin{figure}[!t]
    \centering
    \includegraphics[scale=0.4]{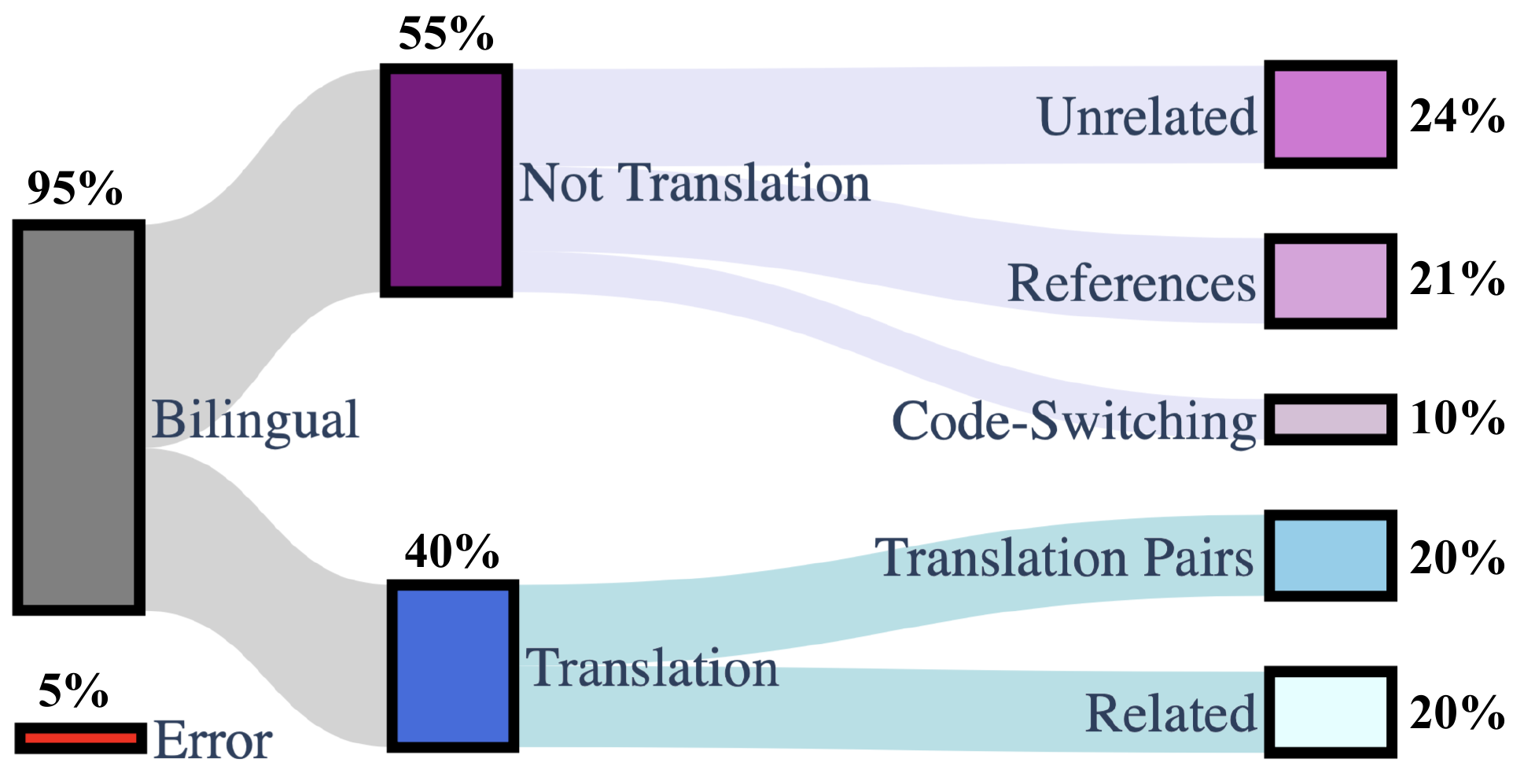}
    \caption{Typology of bilingual instances, along with their distribution within an \textsc{en-fr} annotated sample. Bilingual instances cover a range of cross-lingual phenomena, including cases of translated content.}
    \label{fig:typology}
\end{figure}

\paragraph{Findings} Our qualitative analysis of a sample of $100$ English-French bilingual instances  reveals that bilingualism manifests in various cross-lingual phenomena (examples of bilingual instances are presented in Table~\ref{tab:bilingual_examples} of Appendix~\ref{sec:appendix}).
%
Our detection approach is reasonably accurate: only $5\%$ of instances correspond to errors mostly attributed to language identification issues (i.e., the detected instances are indeed bilingual, but at least one of the two languages is not English or French).
Each correctly detected bilingual instance is annotated as belonging
to one of five categories, with the typology shown in Figure~\ref{fig:typology}. 

Most bilingual instances ($55\%$) fall under the broader class of ``Not Translations'' and cover cases where the two languages encode information that does not correspond to translation content. 
This class is further decomposed into three sub-classes. First, we found a few instances ($10\%$) of code-switching where one or two speakers alternate between two languages in the context of a single conversation. As expected, most code-switching instances were spotted in social media conversations, as it is primarily used within multilingual communities in informal communication. Second, we observed that many bilingual instances ($21\%$) are attributed to references, where named entities or bibliography entries are cited in their native language, such as instances drawn from Wikipedia. Third, we also found a considerable number of bilingual instances ($24\%$) that include completely unrelated content in the two languages that just happened to co-exist within the same web page. 

The remaining bilingual instances are evenly distributed ($20\%$) across two categories that fall loosely under the rubric of ``Translations''. Here, we distinguish between cases where some amount of the text expresses a typical translation relation and cases where content across languages is semantically related, but not exactly by translation. 
The latter involves a rich spectrum of cross-lingual semantic relations, including cross-lingual entailment, summarization, and paraphrasing, mainly noticed within books in the genre of literary criticism and interpretation. We also spotted a few cases of forum discussions around explanations of translation or stylistic manipulation of translations.

\subsection{Detecting Translation Pairs}
\label{sec:quant3}

Our manual analysis exposed an opportunity to automatically extract and count translated sentence pairs (\textit{translation pairs} for short).
We cast the problem of within-instance translation detection as a local mining task following recent advances in parallel text acquisition. Concretely, for each bilingual instance from \S\ref{sec:quant1}, we run a sentence breaker and extract two pools of candidate sentences $x$ and $y$ in the two languages. The language of each sentence is inferred by majority voting over token-level language tags.
%
Whichever language has fewer sentences is labeled the embedded language and the other becomes the primary. Each candidate sentence is then encoded to a vector representation using the \textsc{labse}~\cite{feng-etal-2022-language} cross-lingual sentence encoder. Translation pairs are extracted by finding the most similar primary sentence for each embedded sentence and then checking whether the cosine distance of their representations falls below a threshold. We choose a threshold of $0.6$ on the cosine distance to mine plausible translation pairs, following \citet{feng-etal-2022-language}.
%
We also apply a series of length-and-language-based heuristic data quality filters, adapted from Alibaba's WMT Data Filtering submissions~\cite{lu-etal-2018-alibaba,lu-etal-2020-alibaba}, described in Appendix~\ref{sec:appendix_filters}. 

Note that this extraction process is oblivious to document structure: the instance may be formatted as parallel sentences, paragraphs, documents, or as a free-form discussion that happens to mention both a sentence and its translation. 
Our extraction is also incapable of detecting translation relations below the sentence level.
If we can extract at least one translation pair from an instance, then we label it as a \textit{translation instance}.

\begin{figure}[!t]
    \centering
    \includegraphics[scale=0.2]{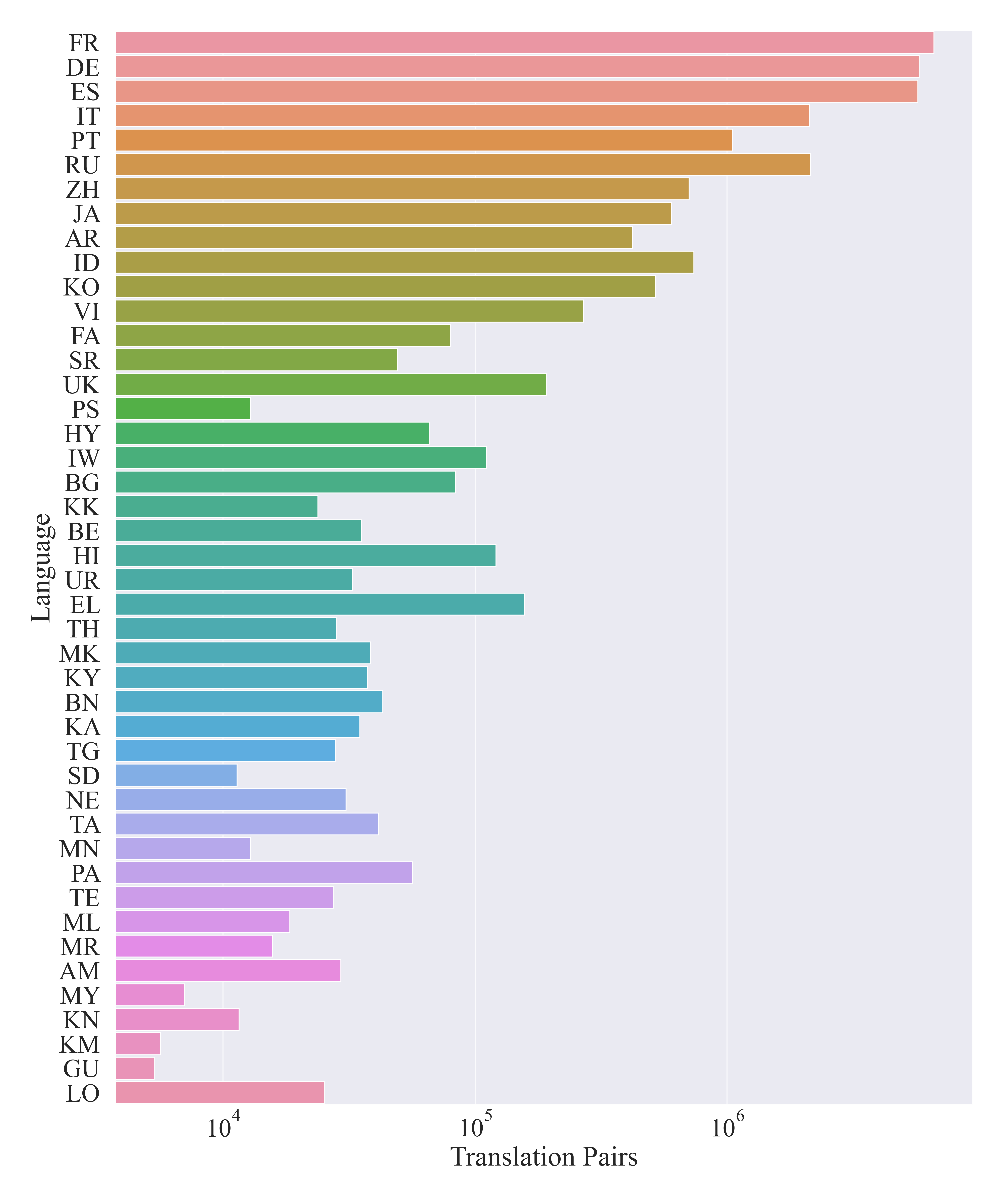}
    \caption{Number of mined translation pairs within PaLM's training instances. PaLM consumes thousands of translation pairs across (at least) $44$ languages.}
    \label{fig:parallel_stats}
\end{figure}

\begin{figure}[!t]
    \centering
    \subfloat[\centering $r=0.944$]{{\includegraphics[width=5.65cm]{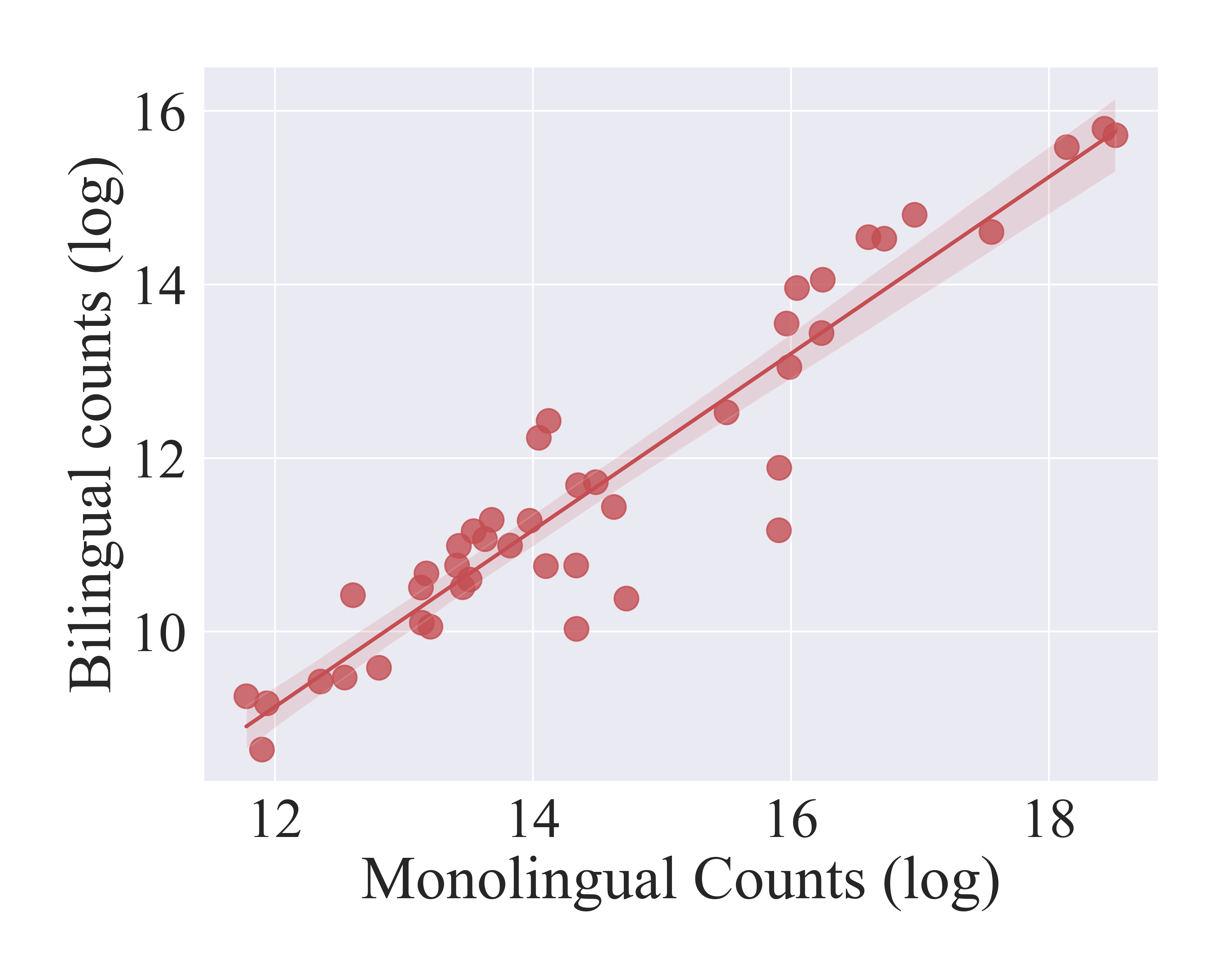} }}%
    \hfill
    \subfloat[\centering $r=0.938$ ]{{\includegraphics[width=5.65cm]{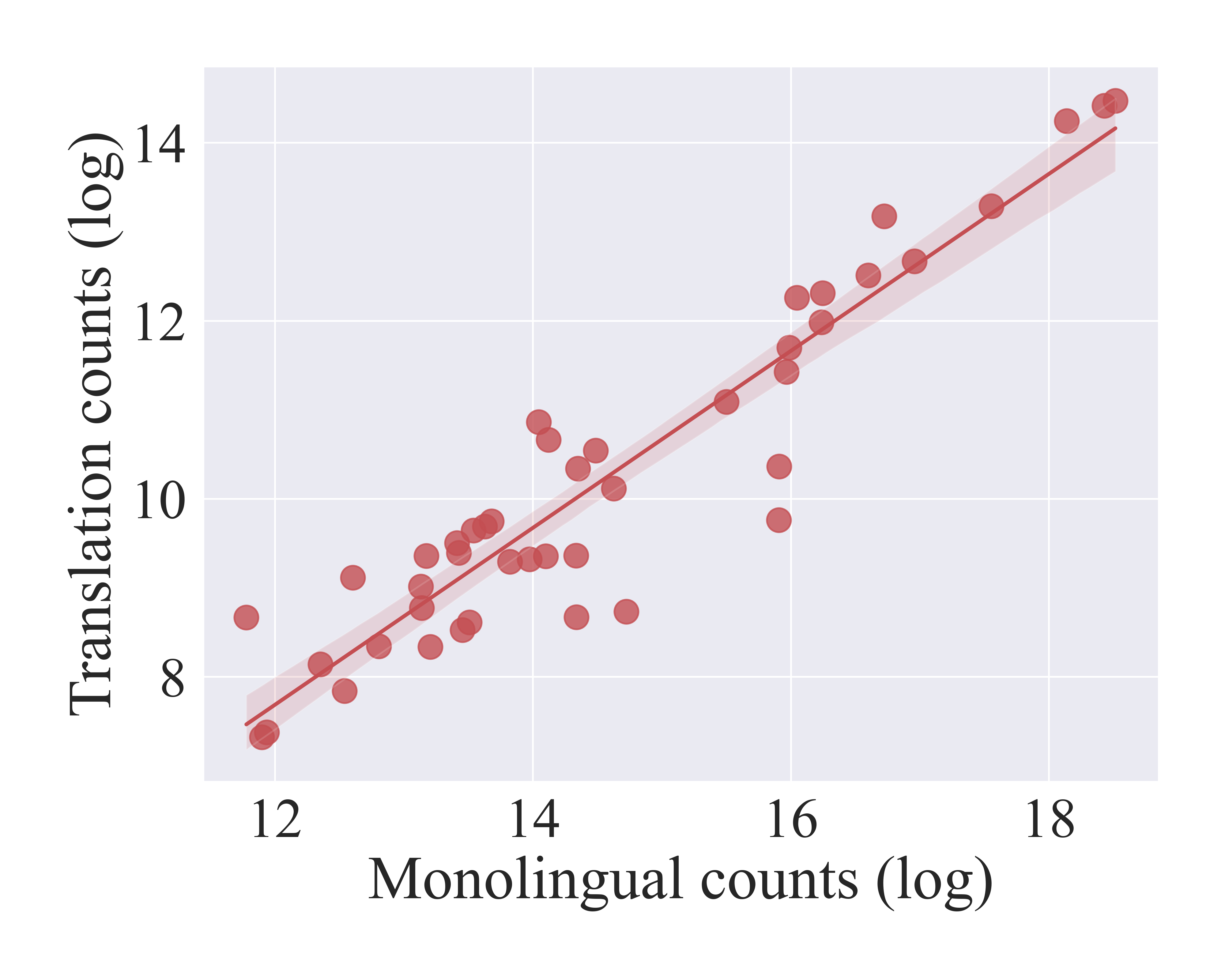} }}%
    \caption{Pearson correlations between counts of monolingual instances with (a) bilingual and (b) translation instances. The number of bilingual and translation instances correlates strongly with the number of monolingual instances.}%
    \label{fig:data_correlation}
\end{figure}

\paragraph{Findings}
We find that $0.34\%$ of PaLM's training instances contain at least one translation pair.
Note that this number provides a lower bound on the amount of incidental bilingualism and translation that PaLM consumes, as we are restricted to a specific set of language pairs, and we only study bilingualism with English.
Figure~\ref{fig:parallel_stats} presents the number of translation pairs we mined within PaLM's training instances between  English  and each language. At a minimum, PaLM consumes thousands of parallel texts for all language pairs studied, while for high-resource languages it sees more than a million translation pairs.

Furthermore, we investigate the correlation between the number of monolingual instances in each language and their bilingual and translation counterparts. Our results in Figure~\ref{fig:data_correlation} indicate that, surprisingly, the monolingual counts in each language correlate strongly with the bilingual (r=$0.944$) and translation (r=$0.938$) counts. This strong correlation implies that, when working at scale, we can predict the bilingual and translation sizes for a given language (within an error rate) by simply counting monolingual instances.

\subsection{Discovering Natural Prompts}\label{sec:qual4}

After identifying a smaller-scale set consisting of training instances that contain translation pairs, we further manually inspect them to understand how the translation task is naturally modeled by PaLM. 
We find that sentence-level translations are presented within a training instance in three ways. The majority of them appear across paragraphs and do not follow a canonical pattern. Among the remainder, we noticed two canonical patterns: translation pairs that belong to stacked translated paragraphs (e.g., $\{x_1, x_2, y_1, y_2\}$) and interleaved translations where a sentence and each translation are adjacent to each other (e.g., $\{x_1, y_1, x_2, y_2\}$). Among the latter, we saw an opportunity to extract natural prompts automatically.
We do so by analyzing the prefixes of the translation pairs mined in \S\ref{sec:quant3}. Drawing on our manual observations, we mine the most frequent prefixes per language pair that follow a simple colon prompt format: any sequence of non-whitespace characters followed by a colon.
Finally, we manually filter the automatically mined prefix lists to look for consistent natural prompt patterns across languages.

\begin{table}[!t]
    \centering
    \scalebox{0.85}{
    \begin{tabular}{lrrrr}
    \rowcolor{gray!50}
    & \textbf{Default} & \textbf{Code} & \textbf{Native} & \textbf{Translation} \\
    \textbf{\textsc{high}} & $1{,}207$ & $506$	& $781$ & $831$  \\
    \textbf{\textsc{medium}} &	$219$  & $62$ &	$136$ &	$352$\\
    \textbf{\textsc{low}} & $38$ &$0$ & $64$ & $122$ \\
    \textbf{\textsc{all}} & $1{,}464$ & $568$ &	$981$ &	$1{,}305$  \\
    \end{tabular}}
    \caption{Data-driven prompt counts within PaLM's translation pairs,  grouped by resourcedness.}
    \label{tab:prompts_grouped_counts}
\end{table}
\input{tables/prompts_mt}

\paragraph{Findings} Table~\ref{tab:prompts_grouped_counts} presents the results of our prompt discovery module followed by manual filtering to extract plausible translation prefixes. First, we found empirically that one of the most frequent translation prompts that naturally arises in the data is the \textbf{default} prompt adopted by most \textsc{mt} research with \textsc{llm}s: source and target language names in English followed by a colon (e.g., ``French:''). We also found three alternative prompts that are frequently presented within incidental translation pairs:
\begin{inparaenum}[i)]
\item \textbf{code}: source and target \textsc{iso} language codes (e.g., ``\textsc{fr}:''), 
\item \textbf{native}: source and target language names in their respective languages (e.g., ``Français:''),
\item \textbf{translation}: source language in English, and the word ``translation'' in the target language (e.g., ``Traduction:'').
\end{inparaenum}
Interestingly, prompt types are not evenly distributed across our language groups: language codes appear primarily with high-resource languages, while low-resource languages favor prompts written in their native language.
We include a complete list of prompt counts per language in Figure~\ref{fig:prompt_discovery_full} of Appendix~\ref{sec:appendix}.

%% file: tables/prompts_mt.tex
\begin{table*}[!t]
    \centering
    \scalebox{0.54}{
    \begin{tabular}{l|rr@{\hskip 0.3in}|rrr@{\hskip 0.3in}|rrr@{\hskip 0.3in}|rrr@{\hskip 0.3in}||rr@{\hskip 0.3in}|rrr}
        
    \rowcolor{gray!50}
     & \multicolumn{2}{c}{\textbf{Default} (\textit{zero})} & \multicolumn{3}{c}{\textbf{Code} (\textit{zero})} & \multicolumn{3}{c}{\textbf{Native} (\textit{zero})} & \multicolumn{3}{c}{\textbf{Translation} (\textit{zero})} & \multicolumn{2}{c}{\textbf{Default} (\textit{few})} & \multicolumn{3}{c}{\textbf{Native} (\textit{few})}\\

    \rowcolor{gray!5}
     & \multicolumn{1}{c}{\textsc{qual.}} & \multicolumn{1}{c}{\textsc{lang.}\%} & \multicolumn{1}{c}{\textsc{qual.}} & \multicolumn{1}{c}{$\delta$} & \textsc{lang.}\% &  \multicolumn{1}{c}{\textsc{qual.}} & \multicolumn{1}{c}{$\delta$} & \multicolumn{1}{c}{\textsc{lang.}\%} & \multicolumn{1}{c}{\textsc{qual.}} & \multicolumn{1}{c}{$\delta$} & \multicolumn{1}{c}{\textsc{lang.}\%} & \multicolumn{1}{c}{\textsc{qual.}} & \multicolumn{1}{c}{\textsc{lang.}\%} &  \multicolumn{1}{c}{\textsc{qual.}} & \multicolumn{1}{c}{$\delta$} & \multicolumn{1}{c}{\textsc{lang.}\%}\\
    
    \rowcolor{gray!20}
    \multicolumn{17}{c}{\textsc{en}$\rightarrow$\textsc{xx}} \\
     \textsc{\textbf{high}} &  $52.8$ & \textcolor{gray}{$81.5$} & $56.7$ & \grpos{4.0}{-0.1}{30} & \textcolor{gray}{$89.7$} & $60.8$ & \grpos{8.0}{-0.1}{30} & \textcolor{gray}{$99.5$} & $59.1$ & \grpos{6.3}{-0.1}{30} & \textcolor{gray}{$96.2$} & $62.9$ & \textcolor{gray}{$99.7$} & $63.1$ & \grpos{0.2}{-0.1}{30} & \textcolor{gray}{$99.7$} \\
     \textsc{\textbf{medium}} &  $30.6$ & \textcolor{gray}{$64.8$} & $17.2$ & \grneg{-13.4}{-50}{0.1} & \textcolor{gray}{$33.4$} & $46.1$ & \grpos{15.5}{-0.1}{30} & \textcolor{gray}{$92.8$} & $44.6$ & \grpos{14.0}{-0.1}{30} & \textcolor{gray}{$81.7$} & $53.4$ & \textcolor{gray}{$99.7$} & $53.4$ & \grneg{-0.0}{-50}{0.1} & \textcolor{gray}{$99.7$} \\
     \textsc{\textbf{low}} &  $28.3$ & \textcolor{gray}{$69.0$} & $2.7$ & \grneg{-25.6}{-50}{0.1} & \textcolor{gray}{$3.4$} & $42.3$ & \grpos{14.0}{-0.1}{30} & \textcolor{gray}{$98.6$} & $38.1$ & \grpos{9.8}{-0.1}{30} & \textcolor{gray}{$82.4$} & $47.4$ & \textcolor{gray}{$100.0$} & $47.4$ & \grpos{0.0}{-0.1}{30} & \textcolor{gray}{$100.0$} \\
     \textsc{\textbf{all}} &  $31.1$ & \textcolor{gray}{$69.1$} & $11.2$ & \grneg{-19.9}{-50}{0.1} & \textcolor{gray}{$18.8$} & $45.0$ & \grpos{13.8}{-0.1}{30} & \textcolor{gray}{$97.2$} & $41.6$ & \grpos{10.5}{-0.1}{30} & \textcolor{gray}{$83.5$} & $50.3$ & \textcolor{gray}{$99.9$} & $50.3$ & \grpos{0.0}{-0.1}{30} & \textcolor{gray}{$99.9$} \\

    \addlinespace[0.1cm]
    
    \rowcolor{gray!20}
    \multicolumn{17}{c}{\textsc{xx}$\rightarrow$\textsc{en}} \\    

     \textsc{\textbf{high}} &  $37.6$ & \textcolor{gray}{$99.7$} & $38.5$ & \grpos{0.9}{-0.1}{30} & \textcolor{gray}{$99.6$} & $37.7$ & \grpos{0.1}{-0.1}{30} & \textcolor{gray}{$99.7$} & $35.4$ & \grneg{-2.2}{-50}{0.1} & \textcolor{gray}{$99.1$} & $40.6$ & \textcolor{gray}{$99.7$} & $40.8$ & \grpos{0.2}{-0.1}{30} & \textcolor{gray}{$99.7$} \\
     \textsc{\textbf{medium}} &  $36.9$ & \textcolor{gray}{$99.5$} & $34.8$ & \grneg{-2.1}{-50}{0.1} & \textcolor{gray}{$94.0$} & $36.6$ & \grneg{-0.3}{-50}{0.1} & \textcolor{gray}{$99.1$} & $35.1$ & \grneg{-1.8}{-50}{0.1} & \textcolor{gray}{$95.7$} & $40.0$ & \textcolor{gray}{$99.6$} & $40.0$ & \grpos{0.2}{-0.1}{30}  & \textcolor{gray}{$99.6$} \\
     \textsc{\textbf{low}} &  $30.9$ & \textcolor{gray}{$99.3$} & $28.5$ & \grneg{-2.3}{-50}{0.1} & \textcolor{gray}{$93.7$} & $28.4$ & \grneg{-2.5}{-50}{0.1} & \textcolor{gray}{$98.8$} & $28.8$ & \grneg{-2.1}{-50}{0.1} & \textcolor{gray}{$90.3$} & $35.4$ & \textcolor{gray}{$99.7$} & $35.4$ & \grpos{0.0}{-0.1}{30} &  \textcolor{gray}{$99.6$} \\
     \textsc{\textbf{all}} &  $33.0$ & \textcolor{gray}{$99.4$} & $31.0$ & \grneg{-2.0}{-50}{0.1} & \textcolor{gray}{$94.3$} & $31.3$ & \grneg{-1.7}{-50}{0.1} & \textcolor{gray}{$99.0$} & $31.0$ & \grneg{-2.0}{-50}{0.1} & \textcolor{gray}{$92.4$} & $37.0$ & \textcolor{gray}{$99.7$} & $37.0$ & \grpos{0.0}{-0.1}{30} & \textcolor{gray}{$99.6$} \\

\end{tabular}}
\caption{Comparison of prompt selection on \textsc{flores} devtest, for zero- and few (5)-shot prompting. \textsc{qual.} corresponds to translation quality (chrF for \textsc{en}$\rightarrow$\textsc{xx}, BLEU for \textsc{xx}$\rightarrow$\textsc{en}),
\textsc{lang.}$\%$ represents PaLM's sentence-level accuracy in producing text in the correct target language, and $\delta$ gives the translation quality difference from the ``Default" prompt. Native data-driven prompts improve zero-shot, out-of-English (\textsc{en}$\rightarrow$\textsc{xx}) translation quality largely by guiding PaLM to generate text in the correct target language.}
\label{tab:prompts_grouped}
\end{table*}

%% file: S4.Analyzing.tex
\section{Analyzing the Impact of Bilingualism}\label{sec:analyzing}

We analyze the impact of bilingualism on the translation capabilities of PaLM with a series of \textsc{mt} experiments on the \textsc{flores}-101~\cite{goyal-etal-2022-flores} evaluation set, which provides translations of a common set of English Wikipedia sentences into all of our $44$ languages. We report results on the $1{,}012$ sentence devtest set. We use the $997$ sentence dev set primarily as a source of randomly-drawn exemplars when reporting $5$-shot results. We report \textsc{bleu}~\cite{papineni-etal-2002-bleu} for into-English translation and chrF~\cite{popovic-2015-chrf} for out-of-English translation, both computed by Sacrebleu~\cite{post-2018-call} with default settings. For \textsc{llm}-based translation, we follow the template from \citet{vilar2022prompting} unless stated otherwise:
\newcommand{\niceBrackets}[1]{&\texttt{[}&\omit\hfill $#1$\hfill\texttt{]}}
\newcommand{\tightX}{\hspace{0em}}
\newcommand{\tightY}{\vspace{-0.25em}}
\begin{alignat*}{5}
   \tightX &\texttt{[source]: }& \niceBrackets{X} \tightY \\
   \tightX &\texttt{[target]: } \tightY
\end{alignat*}
where \texttt{[source]}, and \texttt{[target]} are the source and target language names (in English) and  [$X$] is the source text. When present, few-shot exemplars are provided above the template in the same format, as detailed in Appendix~\ref{sec:appendix_prompting_details}.

\subsection{Prompting PaLM with Natural Prompts}

We prompt the original $540$B parameter PaLM model with templates that use naturally-occurring prefixes of incidental translations, as discussed in \S\ref{sec:qual4}. In our template, we replace  \texttt{[source]} and  \texttt{[target]} with each alternative, data-driven prompt. We experiment with zero-shot and 5-shot prompting.

\paragraph{Findings} Table~\ref{tab:prompts_grouped} presents average translation quality results for different prompts across high, medium, and low resource settings. We present the complete, per language results in Table~\ref{tab:prompts_full} of Appendix~\ref{sec:appendix}.
%
When translating into English (\textsc{\textbf{xx$\rightarrow$en}}), the default prompt yields the best results, while alternative prompts result in a small degradation in quality; overall, translating into English seems to be robust across different prompts supported by our data. On the other hand, PaLM's translation quality is surprisingly sensitive to the choice of prompt when translating out of English (\textbf{\textsc{en$\rightarrow$xx}}): simply changing the default prompt to its native variant improves quality by $~14$ chrF points, with most of the improvement reported in medium and low-resource languages. The ``translation'' prompt also yields consistent improvements over the default. Finally, prompting with language codes only improves translation out of English for the high-resource group---this is expected as this prompt was only present for a few high-resource languages.
Further analysis of out-of-English results
reveals that native prompts trigger text in the desired language, while the default prompt results in high rates of generating the wrong target language (see gray percentages in Table~\ref{tab:prompts_grouped}). The output's target language is determined by a sequence-level language-identification tool~\cite{botha-etal-2017-natural}.
%

Finally, although choosing natural prompts that arise from the data can help us better understand PaLM's zero-shot capabilities, large differences between prompts do not carry over to the few-shot setting (right-most columns of Table~\ref{tab:prompts_grouped}).


\subsection{Extrinsic Evaluation of Translation Pairs} \label{sec:reverse_ablation}

It is one thing to report counts of translation pairs mined from bilingual instances, but is the resulting bitext of high quality?
We adopt the parallel text quality evaluation framework of the \textsc{wmt} Shared Task on Parallel Corpus Filtering and Alignment~\cite{koehn-etal-2020-findings} and train supervised neural machine translation models from scratch on the mined translations. This allows us to jointly assess the quality of PaLM's translation content and our extraction heuristics. We focus this analysis on \textsc{fr}$\rightarrow$\textsc{en}, PaLM's highest-resource language pair.

\paragraph{Data} For PaLM translation pairs, we explore a number of thresholds on the \textsc{labse} distance. To put our results in perspective, we additionally train a model on all pairs from the WMT$14$ \textsc{fr}$\rightarrow$\textsc{en} task~\cite{bojar-etal-2014-findings} and on random samples thereof to establish fair data comparison points at notable \textsc{labse} thresholds. Sentence counts for all conditions are shown in Table~\ref{tab:reverse}.

\paragraph{Architecture} We adopt the $6$-layer encoder-decoder Transformer Base~\cite{Vaswani2017Attention} architecture, with minimal hyper-parameter tuning. Shared sentence piece~\cite{kudo-richardson-2018-sentencepiece} vocabularies with $32$K tokens are constructed from bitext for each scenario. Dropout is set to $0.3$ for all systems except for the full \textsc{wmt} system, which uses $0.1$. Systems are trained up to $450$K steps with a batch size of $1{,}024$. Checkpoints are selected by \textsc{flores} dev \textsc{bleu}.

\paragraph{Findings} Table~\ref{tab:reverse} presents the results of our analysis. In general, the mined translation pairs from our analysis pipeline provide useful signal for training supervised \textsc{mt} systems with reasonable translation quality (i.e., ~$37$ to $38$ \textsc{bleu} across various thresholds, compared to $41$ that we achieve using $~40$M translations from available \textsc{wmt} parallel corpora). Moreover, these results confirm that $0.6$ seems to be the right threshold for detecting translation pairs that are useful, or at least not harmful in the presence of other positive signals (i.e., at $0.6$ we are within 1 \textsc{bleu} point of a system trained on the same amounts of \textsc{wmt} parallel text).

\input{tables/reverse_ablation}

\input{tables/ablation}

\subsection{Ablating Incidental Bilingualism}
We now explore the impact of bilingualism on the translation capabilities of PaLM. To do so, we conduct smaller-scale experiments by training $1$B and $8$B parameter models on different training samples to measure the effect of removing various types of multilingual data. 

\paragraph{Architecture}
Our $1$B and $8$B models are scaled-down versions of PaLM with small changes. Like PaLM, each is a decoder-only model trained with a causal language modeling objective, using a dense transformer architecture and a sentence piece tokenizer~\cite{kudo-richardson-2018-sentencepiece} that retains spacing information. 
Unlike PaLM, we do not share key and value tensors across attention heads~\cite{shazeer-2019-fast-transformer}, which should affect only decoding speed. 
We include a hyper-parameter summary in Table~\ref{tab:ablation_hparams} in Appendix~\ref{sec:appendix}.
Also, we use a smaller vocabulary size of $128$K tokens compared to PaLM's $256$K tokens, a concession to fit the models onto available hardware. 
Both $1$B and $8$B train on examples of $2{,}048$ tokens with a batch size of $512$ for $100$K steps. Note that using the same number of examples for both scales means that the $8$B models are likely under-trained; however, holding data quantity constant is useful for directly measuring the effect of model scale.

\paragraph{Data} 
To simulate PaLM's data conditions with smaller models, we begin by partitioning PaLM's training instances into four non-overlapping groups:  \textbf{\textsc{eng}}: English instances, \textbf{\textsc{nen}}: non-English (excluding bilingual) instances, \textbf{\textsc{bil}}: bilingual (excluding translation) instances, and \textbf{\textsc{tra}}: translation instances. We then merge instances within their groups into $2{,}048$ token examples. Counting examples from each group allows us to determine the full data's implicit mixture of these groups: \textbf{\textsc{eng}}: $84.4\%$; \textbf{\textsc{nen}}: $14.1\%$; \textbf{\textsc{bil}}: $1.0\%$; \textbf{\textsc{tra}}: $0.5\%$.
These should not match the instance-level proportions reported earlier, as these count examples, which are merged instances. Also, they will not match the multilinguality proportions reported by \citet{Chowdhery2022PaLMSL}, as we have removed non-natural-language (code) data and any non-English text not in our $44$-language set.
%
We can now sample examples from our partitions to create a smaller training set with the same proportions of incidental bilingualism. No attempt is made to retain PaLM's original proportions for other aspects like data source or language.
Counts for this sample are shown as \textbf{\textsc{full}} in Table~\ref{tab:ablationconditions}. 

We ablate each group in the following order: \textbf{\textsc{tra}}, \textbf{\textsc{bil}} and then \textbf{\textsc{nen}}. At each step, we replace ablated examples with examples from the next group in the chain. The counts for all ablation conditions are shown in Table~\ref{tab:ablationconditions}. 
The \textbf{\textsc{-nen}} setting corresponds to the English-only setting studied by \citet{Blevins2022LanguageCH}, but as they show, this will contain some non-English content due to language-identification errors. Analogous provisos exist for each ablation, as all our automatic tools make errors. We aim to measure the effect of removing most of a type of content, not all of it.

\begin{table}[!t]
    \centering
    \scalebox{0.75}{
    \begin{tabular}{lrrrr}
    \rowcolor{gray!50}
    & \textbf{\textsc{eng}} & \textbf{\textsc{nen}} & \textbf{\textsc{bil}} & \textbf{\textsc{tra}} \\
    \textbf{\textsc{full}} & $43,186,985$ & $7,224,737$ & $517,688$ & $270,590$ \\
    \textbf{\textsc{-tra}} & $43,186,985$ & $7,224,737$ & $788,279$ & \xmark \\
    \textbf{\textsc{-bil}} & $43,186,985$ & $8,013,015$ & \xmark & \xmark \\
    \textbf{\textsc{-nen}}   & $51,200,000$ & \xmark & \xmark & \xmark \\ 
    \end{tabular}}
    \caption{Data statistics for small-scale PaLM ablation experiments in number of $2{,}048$ token examples.}
    \label{tab:ablationconditions}
\end{table}

\paragraph{Findings} Table~\ref{tab:ablation_grouped} presents the results of our ablation---the complete, per language, results are in Table~\ref{tab:full_ablation} of Appendix~\ref{sec:appendix}.
Focusing on our $1$B model, we note that examples containing translation pairs (\textbf{\textsc{tra}}) have an outsized impact on translation quality for being only $0.5\%$ of the training data. In the high-resource \textsc{xx}$\rightarrow$\textsc{en}, zero-shot scenario, replacing \textbf{\textsc{tra}} examples with \textbf{\textsc{bil}} results in a drop of $7.4$ \textsc{bleu}. With \textbf{\textsc{tra}} removed, the additional impact of removing the remaining bilingual instances (\textbf{\textsc{bil}}) is much smaller: $1.2$ \textsc{bleu}. One might expect the utility of translation data to fall off as we add $5$-shot examples at inference time, but \textbf{\textsc{tra}} is still quite important, with its removal resulting in a reduction of $5.9$ \textsc{bleu}. The importance of \textbf{\textsc{tra}} holds throughout our $1$B experiments, to the extent that the system cannot translate at all, i.e. for $5$-shot versions of \textsc{xx}$\rightarrow$\textsc{en} \textsc{medium} and \textsc{en}$\rightarrow$\textsc{xx} \textsc{high}. 

Turning to our $8$B model, we see that translation content continues to have a substantial impact on translation quality, though the absolute score differences have diminished, hovering between $2$-$3$ \textsc{bleu} or $3$-$4$ chrF, depending on the scenario.
This result, where a $4$x increase in parameters leads to a roughly $2$x reduction in the absolute impact of \textbf{\textsc{tra}} suggests that it would be interesting to build scaling laws to study the impact of incidental translation data, which we leave to future work. 
Also, for $5$-shot scenarios, there is no longer such a big difference between the impact of \textbf{\textsc{bil}} and \textbf{\textsc{tra}} data. Given exemplars, the larger model seems to be able to make better use of weaker bilingual signals.

Surprisingly, the $8$B model that does not have access to multilingual content (\textbf{\textsc{-nen}}), exhibits some translation capabilities for \textsc{xx}$\rightarrow$\textsc{en} \textsc{high} (i.e., $17.3$ and $25.9$ \textsc{bleu} for  zero- and few-shot, respectively). A closer look at the per-language breakdown (see Table~\ref{tab:full_ablation}) reveals that those capabilities are restricted to languages written in Latin script. This adds evidence for larger models being better equipped to leverage either sparse signals (i.e., language-identification failures during ablation) and weak signals (i.e., language similarities from shared scripts). As expected, non-English content is critical for translation out of English.

%% file: tables/reverse_ablation.tex
\begin{table}[!t]
    \centering
    \scalebox{0.85}{
    \begin{tabular}{lrr@{\hskip 0.2in}c}
        \rowcolor{gray!50}
        $t$ & \textbf{\textsc{\#translations}} & PaLM (mined) & \textsc{wmt} \\ 
        \textcolor{gray}{\textsc{n/a}} & \textcolor{gray}{$40{,}836{,}876$} & \xmark & \textcolor{gray}{$42.0$} \\
        $0.90$ & $9{,}084{,}429$  & $33.7$ &  \\
        $0.80$ & $7{,}056{,}441$  & $35.7$ &  \\
        $0.70$ & $4{,}874{,}173$  & $36.4$ & \\\rowcolor{gray!10}
        $0.60$ & $3{,}341{,}187$  & $37.3$ &  $38.1$ \\ 
        $0.50$ & $2{,}474{,}703$  & $37.2$ & \\
        $0.40$ & $1{,}948{,}820$  & $37.1$ & \\\rowcolor{gray!10}
        $0.30$ & $1{,}477{,}535$  & $38.4$ & $36.5$ \\       
        $0.20$ & $906{,}937$      & $37.8$ & \\
        $0.15$ & $549{,}705$      & $36.3$ & \\
    \end{tabular}}
    \caption{\textsc{bleu} scores for \textsc{fr}$\rightarrow$\textsc{en} \textsc{nmt} models trained on various translation pairs, evaluated on \textsc{flores} devtest. $t$ corresponds to the \textsc{labse} threshold.
    PaLM-mined translation pairs provide useful signal for training supervised \textsc{nmt} models.}
    \label{tab:reverse}
\end{table}

%% file: tables/ablation.tex
\begin{table*}[!t]
    \centering
    \scalebox{0.58}{
    \begin{tabular}{llrrrr@{\hskip 0.3in}rrrr@{\hskip 0.3in}rrrr@{\hskip 0.3in}rrrr}
        
        \rowcolor{gray!50}
        & & \multicolumn{4}{c}{\textsc{en}$\rightarrow$\textsc{xx} (\textit{0-shot})} & \multicolumn{4}{c}{\textsc{en}$\rightarrow$\textsc{xx} (\textit{5-shot})} & \multicolumn{4}{c}{\textsc{xx}$\rightarrow$\textsc{en} (\textit{0-shot})} & \multicolumn{4}{c}{\textsc{xx}$\rightarrow$\textsc{en} (\textit{5-shot})} \\
        
        \rowcolor{gray!20}
        & & \multicolumn{1}{c}{\textsc{full}} & \multicolumn{1}{c}{\textsc{-\textsc{tra}}} & \multicolumn{1}{c}{\textsc{-bil}} & \multicolumn{1}{l}{\textsc{-nen}}
        & \multicolumn{1}{c}{\textsc{full}} & \multicolumn{1}{c}{\textsc{-\textsc{tra}}} & \multicolumn{1}{c}{\textsc{-bil}} & \multicolumn{1}{l}{\textsc{-nen}}
        & \multicolumn{1}{c}{\textsc{full}} & \multicolumn{1}{c}{\textsc{-\textsc{tra}}} & \multicolumn{1}{c}{\textsc{-bil}} & \multicolumn{1}{l}{\textsc{-nen}}
        & \multicolumn{1}{c}{\textsc{full}} & \multicolumn{1}{c}{\textsc{-\textsc{tra}}} & \multicolumn{1}{c}{\textsc{-bil}} & \multicolumn{1}{c}{\textsc{-nen}}\\
        
        \addlinespace[0.1cm]

    & \textbf{\textsc{high}} &  \gradientchrf{15.7} & \gradientchrf{16.4} & \gradientchrf{15.6} & \gradientchrf{15.1} & \gradientchrf{30.9} & \gradientchrf{18.7} & \gradientchrf{15.8} & \gradientchrf{8.0} & \gradient{12.5} & \gradient{5.1} & \gradient{3.9} & \gradient{1.1} & \gradient{14.8} & \gradient{8.9} & \gradient{6.1} & \gradient{6.1}   \\
    & \textbf{\textsc{medium}} &  \gradientchrf{3.8} & \gradientchrf{4.6} & \gradientchrf{3.6} & \gradientchrf{3.7} & \gradientchrf{11.3} & \gradientchrf{8.1} & \gradientchrf{6.9} & \gradientchrf{3.2} & \gradient{2.9} & \gradient{0.8} & \gradient{1.0} & \gradient{0.2} & \gradient{5.7} & \gradient{2.1} & \gradient{1.7} & \gradient{1.7}  \\
    & \textbf{\textsc{low}} &  \gradientchrf{0.6} & \gradientchrf{0.6} & \gradientchrf{0.5} & \gradientchrf{0.5} & \gradientchrf{6.3} & \gradientchrf{6.7} & \gradientchrf{5.6} & \gradientchrf{3.4} & \gradient{0.3} & \gradient{0.3} & \gradient{0.3} & \gradient{0.1} & \gradient{0.8} & \gradient{0.5} & \gradient{0.2} & \gradient{0.2}   \\ \rot{\rlap{~S=1B}}  
    & \textbf{\textsc{all}} &  \gradientchrf{2.8} & \gradientchrf{3.0} & \gradientchrf{2.7} & \gradientchrf{2.6} & \gradientchrf{9.8} & \gradientchrf{8.2} & \gradientchrf{6.9} & \gradientchrf{3.8} & \gradient{2.1} & \gradient{0.8} & \gradient{0.8} & \gradient{0.2} & \gradient{3.3} & \gradient{1.6} & \gradient{1.1} & \gradient{1.1}   \\
            
        \addlinespace[0.2cm]
    
    & \textbf{\textsc{high}} &  \gradientchrf{21.5} & \gradientchrf{17.7} & \gradientchrf{20.4} & \gradientchrf{17.9} & \gradientchrf{47.7} & \gradientchrf{44.7} & \gradientchrf{40.7} & \gradientchrf{25.8} & \gradient{24.0} & \gradient{22.2} & \gradient{22.4} & \gradient{17.3} & \gradient{30.4} & \gradient{27.4} & \gradient{25.9} & \gradient{25.9}  \\
    & \textbf{\textsc{medium}} &  \gradientchrf{5.1} & \gradientchrf{4.6} & \gradientchrf{5.3} & \gradientchrf{4.7} & \gradientchrf{26.5} & \gradientchrf{23.6} & \gradientchrf{20.3} & \gradientchrf{4.9} & \gradient{13.0} & \gradient{10.2} & \gradient{11.9} & \gradient{4.7} & \gradient{21.4} & \gradient{18.7} & \gradient{16.3} & \gradient{16.3}  \\
    & \textbf{\textsc{low}} &  \gradientchrf{1.2} & \gradientchrf{0.7} & \gradientchrf{1.1} & \gradientchrf{0.8} & \gradientchrf{8.8} & \gradientchrf{8.3} & \gradientchrf{7.4} & \gradientchrf{2.2} & \gradient{2.6} & \gradient{2.0} & \gradient{2.9} & \gradient{0.4} & \gradient{6.6} & \gradient{5.0} & \gradient{4.7} & \gradient{4.7}  \\ \rot{\rlap{~S=8B}}
    & \textbf{\textsc{all}} &  \gradientchrf{4.0} & \gradientchrf{3.2} & \gradientchrf{3.9} & \gradientchrf{3.3} & \gradientchrf{16.8} & \gradientchrf{15.5} & \gradientchrf{13.6} & \gradientchrf{5.1} & \gradient{7.2} & \gradient{5.9} & \gradient{6.9} & \gradient{3.0} & \gradient{12.4} & \gradient{10.5} & \gradient{9.5} & \gradient{9.5}   \\

    \end{tabular}}
    \caption{Translation results on the \textsc{flores} devtest for small-scale PaLM models trained on various ablation conditions. \textsc{en}$\rightarrow$\textsc{xx} translation quality is measured by chrF and \textsc{xx}$\rightarrow$\textsc{en} by \textsc{bleu}.
    Ablating translation pairs (-\textsc{tra}) has a significant impact on the translation capabilities of S=$1$B ($5$-shot) for \textsc{high} resource pairs; this impact decreases with scale (i.e., S=$8$B model).}\label{tab:ablation_grouped}
\end{table*}

%% file: S5.Appendix.tex
\section{Units of Analysis of Training text}\label{sec:appendix_analysis}
Throughout this paper we have adopted special meanings for the common (often interchangeable) terms \textit{document}, \textit{example} and \textit{instance}. Here we make those terms concrete and justify our use of the \textit{instance} as our primary unit of analysis.

\paragraph{Document} A document is a logical unit of text from one of our source corpora: a web page or wiki page from a web-crawled corpus, a conversation from a chat or forum corpus, or a book from a books corpus.

\paragraph{Example} Each PaLM training example is exactly 2,048 subword tokens. These are assembled by concatenating and/or splitting documents to the appropriate length. As such, an example may contain several short documents, and a long document may be spread over several examples. Multiple documents concatenated into a single example are separated by special document-boundary tokens. The relevant features of examples that make them more useful for analysis than documents are:
\begin{itemize}
    \item We know exactly which examples PaLM saw during training.
    \item Examples reflect when co-occurring textual information (for example, a translation pair) was lost due to a document being split into multiple examples.
\end{itemize}
However, examples can also introduce spurious co-occurrences from merged documents. We assume that a language model can and will ignore any merge-induced co-occurrences due to the presence of document separator tokens; therefore, we should ignore them as well. This leads us to our next and final unit.

\paragraph{Instance} Instances are created by splitting examples according to document-separator tokens. Therefore, each instance is either a complete document or a fragment of a single document, and is up to 2,048 tokens in length. Instances have all of the advantages of examples, without introducing spurious co-occurrences, hence why they are our primary unit of analysis.

\section{Bilingual Detection Pipeline Details}\label{sec:appendix_cmx}

\paragraph{CodeMixer Model Details} We use the \textsc{cmx}  (CodeMixer) model~\cite{zhang-etal-2018-fast-compact}---a token-level language identification model, to detect bilingual instances. \textsc{cmx} is a simple feed-forward model that takes as input a set of character and word-level features and
produces a distribution over a set of languages for each token. The entire sequence of language tags is obtained using constrained decoding over a pre-defined set of permitted languages.
The model is trained on a combination of synthetic and real-world translation data (both monolingual and code-mixed with English) for $100$ languages. 
Note that \textsc{cmx} predicts code-mixing between a \textit{pair} of languages, as a result, it does not reliably predict language tags for multilingual instances involving more than two languages. For example, if an instance actually contains English, French, and German text, with German being the least frequent, it will be tagged as containing only English and French; all German words will be mislabeled as one of the other two languages or as ``undefined.''

\paragraph{Algorithmic Description of Bilingual Detection} 
Given a training instance $t=\{t_i\}_{i=1}^n$, a focus set $\mathcal{L}$ of the $44$ studied languages, and a threshold $N$, we detect bilingual instances based on the following steps: 
\begin{inparaenum}[(i)]
\item We start by extracting a sequence of language tags, using the \textsc{cmx} model. 
\item We mark the most frequent language as the primary language, and the other (if exists) as the embedded.
\item If the primary and the embedded languages do not fall under our focus set $\mathcal{L}$, we exclude it from our analysis.
\item If a training instance contains more than $10\%$ of ``undefined'' predictions (e.g., resulting from non-linguistic 
content), it is not annotated as bilingual.
\item Finally, if a training instance contains at least two contiguous segments---consisting of at least $N$ consecutive identical language tags---in different languages, it is annotated as bilingual.
\end{inparaenum}

Given that the \textsc{cmx} model is known to overpredict English tags, we employ a stricter threshold on defining contiguous segments for English ($N=10$) compared to the rest of the languages ($N=5$). For all languages we operate at the token-level, with the exception of Chinese, Japanese, and Korean for which we apply the above algorithm at the character-level. 

\section{Heuristic Translation Pair Filters} \label{sec:appendix_filters}
When extracting translation pairs found within a bilingual instance, our primary quality signal is from the cosine distance between cross-lingual LABSE sentence embeddings. However, we also apply a suite of heuristic filters which help catch non-translations that slip through this primary filter. These filters are adapted from Alibaba's WMT Data Filtering submissions~\cite{lu-etal-2018-alibaba,lu-etal-2020-alibaba}. When a tokenization is required for token counts or edit distance, we use tokens from the mBERT tokenizer~\cite{devlin-etal-2019-bert}. The filters are as follows:
\begin{inparaenum}
    \item both sentences must respect a min ($3$) and max ($200$) token length;
    \item we enforce a max length ratio ($2$x) between sentences;
    \item we enforce a min edit distance ($2$) and a min edit distance ratio ($0.1$) between sentences; 
    \item we apply a secondary, sequence-level language-identification tool~\cite{botha-etal-2017-natural} to re-identify each side of the pair and ensure that the two halves are written in different languages.
\end{inparaenum}
When extracting sentences to train Transformer Base MT systems in \S\ref{sec:reverse_ablation}, the different-language check is replaced by a check to ensure that the translation pair respects the language pair being studied, i.e.: one sentence is in English and the other is in French.

\section{Prompting Details}\label{sec:appendix_prompting_details}

For $5$-shot prompting experiments we used the following format (e.g., for French to English translation):
\begin{alignat*}{5}
   \tightX &\texttt{French: }& \niceBrackets{X_1} \tightY \\
   \tightX &\texttt{English: } & \niceBrackets{Y_1} \tightY \\
   & ... \\
   \tightX &\texttt{French: }& \niceBrackets{X_5} \tightY \\
   \tightX &\texttt{English: } & \niceBrackets{Y_5} \tightY \\   
   \tightX &\texttt{French: }& \niceBrackets{X} \tightY \\
   \tightX &\texttt{English: } \tightY \\   
\end{alignat*}
Each slot ($X_i$, $Y_i$) is filled with five translation examples that are randomly sampled from the devtest split of the FLORES dataset, while the final slot $X$, is filled with the source text that comes from the test split of FLORES.

\section{Additional Tables and Figures}\label{sec:appendix}
\begin{table*}[!h]
    \centering
    \scalebox{0.8}{
\begin{tabular}{lrrrrrl}
    \rowcolor{gray!50}
    & \textsc{\#layers} & \textsc{\#heads} & \textsc{dimension} & \textsc{data size} & \textsc{computation}\\
    \textbf{\textsc{1B}}   & $16$ & $8$ & $2{,}048$ & $0.1$T & 128 TPUv3 chips for 3 days\\ 
    \textbf{\textsc{8B}} & $32$ & $16$ & $4{,}096$ & $0.1$T & 512 TPUv3 chips for 5 days\\
    \textbf{\textsc{Palm}} & $118$ & $48$ & $18{,}432$ & $2.0$T & See \citet{Chowdhery2022PaLMSL} \\
    \end{tabular}}
    \caption{Ablation hyper-parameters. \textsc{feed-forward dimension} is always \textsc{dimension} times 4. Training data size is measured in trillions (T) of subword tokens.}
    \label{tab:ablation_hparams}
\end{table*}

\input{tables/full_stats}

\input{tables/bilingual_examples}

\begin{figure*}[!ht]
    \centering
    \includegraphics[scale=0.75]{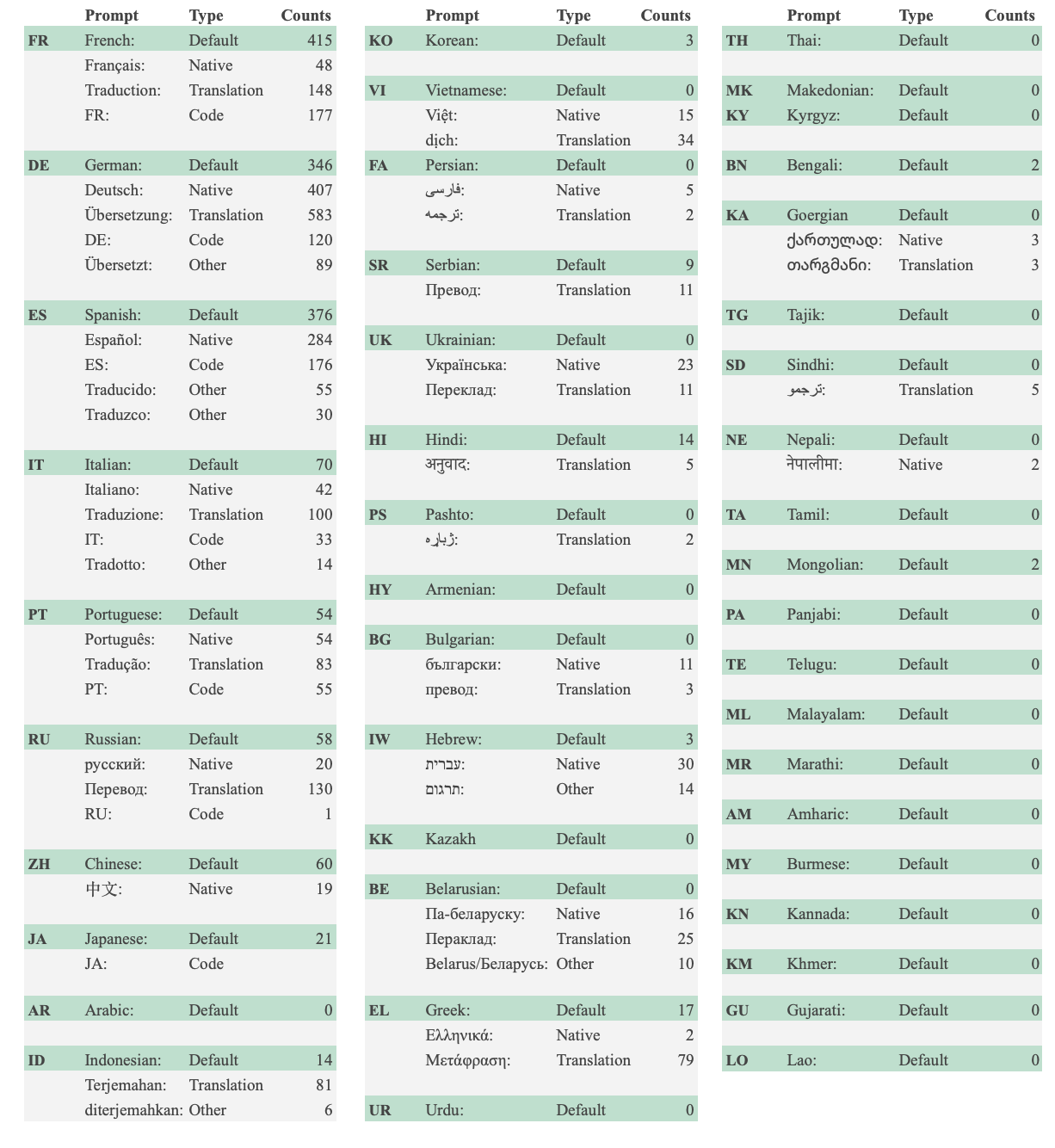}
    \caption{Data-driven prompt counts within PaLM's translation pairs across $44$ languages.}
    \label{fig:prompt_discovery_full}
\end{figure*}

\input{tables/full_prompts}

\input{tables/full_ablation}

\input{tables/per_source_counts}

%% file: tables/full_stats.tex
\begin{table*}[!ht]
    \centering
    \scalebox{0.9}{
    \begin{tabular}{lcrrrr}
        \rowcolor{gray!50} 
        \textsc{\textbf{language}} & \textsc{\textbf{iso}} & \textsc{\textbf{monolingual}} & \textsc{\textbf{bilingual}} & \textsc{\textbf{translation}} & \textsc{\textbf{parallel texts}} \\   
 
        English   &  \textsc{en} & $2{,}086{,}622{,}555{,}000$ &  &  &  \\\rowcolor{gray!10}
        French    &\textsc{fr} & $109{,}994{,}921$ & $6{,}743{,}637$ & $1{,}929{,}032$ & $6{,}618{,}381$ \\
        German    &\textsc{de} & $100{,}952{,}945$ & $7{,}258{,}561$ & $1{,}826{,}701$ & $5{,}780{,}856$ \\\rowcolor{gray!10}
        Spanish    &\textsc{es} & $75{,}311{,}571$ & $5{,}860{,}634$ & $1{,}538{,}549$ & $5{,}717{,}352$ \\
        Italian    &\textsc{it} & $42{,}071{,}597$ & $2{,}204{,}919$ & $591{,}329$ & $2{,}128{,}730$ \\\rowcolor{gray!10}
        Portuguese  &  \textsc{pt} & $23{,}175{,}895$ & $2{,}685{,}160$ & $317{,}735$ & $1{,}048{,}717$ \\
        Russian    &\textsc{ru} & $18{,}307{,}304$ & $2{,}045{,}770$ & $527{,}159$ & $2{,}142{,}065$ \\
\end{tabular}}
    \caption{Numbers of monolingual, bilingual, and translation instances across the $44$ languages studied.}
    \label{tab:full_stats}
\end{table*}

%% file: tables/bilingual_examples.tex
\begin{table*}[!t]
    \scalebox{0.85}{
    \centering
    \begin{tabular}{p{4cm}p{14cm}}

    \rowcolor{gray!20}  \multicolumn{2}{c}{\textbf{\textsc{not translation signal}}}
    \\
    
    \addlinespace[0.4em]
     Code-Switching    &
     \textit{\textcolor{purple!85}{Voilà j'ai un problème avec certaines cinématiques du jeu. Je ne peux pas voir l'introduction ni les présentations de races par contre je peux voir les présentations de classes... Si quelqu'un pouvait m'aider ce serait sympa. Merci d'avance.}  I can understand french only a bit... Can you see this folder and if yes is anything into this folder? \textcolor{purple!85}{J'ai bien un dossier raw/fr mais à l'intérieur il n'y a pas introcinematic.}  Well, could take a look into the folder "raw/en" or/and "raw/de", is there a folder called "introcinematic"?  \textcolor{purple!85}{Dans raw/de je n'ai rien non plus mais dans raw/en j'ai bien le dossier.}}\\
     
    \addlinespace[0.4em]
    \cmidrule{2-2}
    \addlinespace[0.4em]
     
     References & 
     \textit{Lagrange derives the integrals of momentum, moment of momentum, and energy, use of special properties of the potential function tends to conceal their meanings. For three bodies, the results are given in § II of his \textcolor{purple!85}{"Essai sur le problcme des trois corps," Prix de Vacad.} sci. Paris  Finally, the principle of virtual work for dynamics, on which the entire \textcolor{purple!85}{Micbanique Analitique} is founded, had been given more than twenty years earlier in §IV of his \textcolor{purple!85}{"Recherchcs sur la libration de la lune, dans lesquelles on tache dc rcsoudre la question proposce par l'Academie royale des sciences pour le prix de 1'annee $1764$," Prix de Vacad.} sci. Paris $9$, $1764$---Euvres $6$, $5-61$).}\\
\end{tabular}}
    \caption{Examples of bilingual instances detected within PaLM training data.}
    \label{tab:bilingual_examples}
\end{table*}

%% file: tables/full_prompts.tex
\begin{table*}[!ht]
    \centering
    \scalebox{0.5}{
    \begin{tabular}{ll|rr@{\hskip 0.3in}|rrr@{\hskip 0.3in}|rrr@{\hskip 0.3in}|rrr@{\hskip 0.3in}||rr@{\hskip 0.3in}|rrr}
        
    \rowcolor{gray!50}
    & & \multicolumn{2}{c}{\textbf{Default} (\textit{zero})} & \multicolumn{3}{c}{\textbf{Code} (\textit{zero})} & \multicolumn{3}{c}{\textbf{Native} (\textit{zero})} & \multicolumn{3}{c}{\textbf{Translation} (\textit{zero})} & \multicolumn{2}{c}{\textbf{Default} (\textit{few})} & \multicolumn{3}{c}{\textbf{Native} (\textit{few})}\\

    \rowcolor{gray!5}
    & & \multicolumn{1}{c}{\textsc{qual.}} & \multicolumn{1}{c}{\textsc{lang.}\%} & \multicolumn{1}{c}{\textsc{qual.}} & \multicolumn{1}{c}{$\delta$} & \textsc{lang.}\% &  \multicolumn{1}{c}{\textsc{qual.}} & \multicolumn{1}{c}{$\delta$} & \multicolumn{1}{c}{\textsc{lang.}\%} & \multicolumn{1}{c}{\textsc{qual.}} & \multicolumn{1}{c}{$\delta$} & \multicolumn{1}{c}{\textsc{lang.}\%} & \multicolumn{1}{c}{\textsc{qual.}} & \multicolumn{1}{c}{\textsc{lang.}\%} &  \multicolumn{1}{c}{\textsc{qual.}} & \multicolumn{1}{c}{$\delta$} & \multicolumn{1}{c}{\textsc{lang.}\%}\\

    \rowcolor{gray!20}
    \multicolumn{18}{c}{\textsc{en}$\rightarrow$\textsc{xx}} \\
        
      & \textsc{\textbf{fr}} &  $57.8$ & \textcolor{gray}{$79.2$} & $63.6$ & \grpos{5.8}{-0.1}{30} & \textcolor{gray}{$90.4$} & $68.1$ & \grpos{10.3}{-0.1}{30} & \textcolor{gray}{$99.5$} & $65.4$ & \grpos{7.5}{-0.1}{30} & \textcolor{gray}{$94.8$} & $70.7$ & \textcolor{gray}{$99.6$} & $70.9$ & \grpos{0.2}{-0.1}{30} & \textcolor{gray}{$99.7$} \\
    & \textsc{\textbf{de}} &  $52.3$ & \textcolor{gray}{$76.7$} & $59.5$ & \grpos{7.2}{-0.1}{30} & \textcolor{gray}{$92.6$} & $63.0$ & \grpos{10.7}{-0.1}{30} & \textcolor{gray}{$99.7$} & $62.2$ & \grpos{9.9}{-0.1}{30} & \textcolor{gray}{$97.8$} & $65.4$ & \textcolor{gray}{$99.9$} & $65.3$ & \grneg{-0.0}{-50}{0.1} & \textcolor{gray}{$99.9$} \\
\end{tabular}}
\caption{Comparison of prompt selection on \textsc{flores} devtest, for zero- and few-shot prompting. \textsc{qual.} corresponds to translation quality (chrF for \textsc{en}$\rightarrow$\textsc{xx} and \textsc{bleu} for \textsc{xx}$\rightarrow$\textsc{en}), \textsc{lang.}$\%$ represents PaLM's accuracy in producing text in the correct target language, and $\delta$ gives the translation quality difference from the ``Default" prompt.}\label{tab:prompts_full}
\end{table*}

%% file: tables/full_ablation.tex
\begin{table*}[!t]
    \centering
    \scalebox{0.5}{
\begin{tabular}{llrrrr@{\hskip 0.3in}rrrr@{\hskip 0.3in}rrrr@{\hskip 0.3in}rrrrc}
        
        \rowcolor{gray!50}
        & & \multicolumn{4}{c}{\textsc{en}$\rightarrow$\textsc{xx} (\textit{0-shot})} & \multicolumn{4}{c}{\textsc{en}$\rightarrow$\textsc{xx} (\textit{5-shot})} & \multicolumn{4}{c}{\textsc{xx}$\rightarrow$\textsc{en} (\textit{0-shot})} & \multicolumn{4}{c}{\textsc{xx}$\rightarrow$\textsc{en} (\textit{5-shot})} & \\
        
        \rowcolor{gray!20}
        & & \multicolumn{1}{c}{\textsc{full}} & \multicolumn{1}{c}{\textsc{-\textsc{tra}}} & \multicolumn{1}{c}{\textsc{-bil}} & \multicolumn{1}{l}{\textsc{-nen}}
        & \multicolumn{1}{c}{\textsc{full}} & \multicolumn{1}{c}{\textsc{-\textsc{tra}}} & \multicolumn{1}{c}{\textsc{-bil}} & \multicolumn{1}{l}{\textsc{-nen}}
        & \multicolumn{1}{c}{\textsc{full}} & \multicolumn{1}{c}{\textsc{-\textsc{tra}}} & \multicolumn{1}{c}{\textsc{-bil}} & \multicolumn{1}{l}{\textsc{-nen}}
        & \multicolumn{1}{c}{\textsc{full}} & \multicolumn{1}{c}{\textsc{-\textsc{tra}}} & \multicolumn{1}{c}{\textsc{-bil}} & \multicolumn{1}{c}{\textsc{-nen}} & Latin Script\\
        
        \addlinespace[0.1cm]
    
    & \textbf{\textsc{fr}} &  \gradientchrf{16.3} & \gradientchrf{19.0} & \gradientchrf{16.8} & \gradientchrf{16.7} & \gradientchrf{35.1} & \gradientchrf{18.7} & \gradientchrf{15.8} & \gradientchrf{10.6} & \gradient{14.9} & \gradient{7.5} & \gradient{6.1} & \gradient{1.8} & \gradient{18.0} & \gradient{12.6} & \gradient{8.3} & \gradient{8.3}  & \cmark \\
    & \textbf{\textsc{de}} &  \gradientchrf{16.4} & \gradientchrf{15.1} & \gradientchrf{14.3} & \gradientchrf{15.4} & \gradientchrf{29.2} & \gradientchrf{18.9} & \gradientchrf{14.9} & \gradientchrf{7.4} & \gradient{15.0} & \gradient{6.2} & \gradient{4.4} & \gradient{1.2} & \gradient{16.4} & \gradient{8.2} & \gradient{6.9} & \gradient{6.9}  & \cmark \\
\end{tabular}}
    \caption{Translation results between $44$ languages and English on \textsc{flores} devtest for small-scale PaLM models. \textsc{en}$\rightarrow$\textsc{xx} results are reported in chrF, and \textsc{xx}$\rightarrow$\textsc{en} results are report in BLEU.}\label{tab:full_ablation}
\end{table*}

%% file: tables/per_source_counts.tex
\begin{table*}[!ht]
    \centering
    \scalebox{0.7}{
    \begin{tabular}{lrrrr}

    \rowcolor{gray!50}
    \textbf{\textsc{source}} & \multicolumn{1}{c}{\textbf{\textsc{en}}} & \multicolumn{1}{c}{\textbf{\textsc{nen}}} & \multicolumn{1}{c}{\textbf{\textsc{bil}}} & \multicolumn{1}{c}{\textbf{\textsc{tra}}} \\

    \rowcolor{gray!20}
    \multicolumn{5}{c}{\textit{Raw counts (tokens)}} \\
    
    Social media conversations (multilingual) $50\%$ &	$756{,}378{,}913{,}006$ &	$169{,}908{,}649{,}039$ &	$6{,}404{,}486{,}427$ &	$1{,}448{,}443{,}476$  \\
    Filtered webpages (multilingual) $27\%$ & $459{,}437{,}466{,}428$ & $38{,}653{,}502{,}458$ & $7{,}387{,}577{,}398$ & $4{,}260{,}754{,}907$ \\
    Wikipedia  (multilingual)  $4\%$ & $12{,}851{,}315{,}601$ & $42{,}010{,}300{,}146$ & $2{,}514{,}892{,}098$ & $1{,}403{,}598{,}754$  \\
    Books (English) $13\%$ & $258{,}396{,}969{,}011$ & $597{,}753{,}715$ & $1{,}605{,}687{,}335$ & $2{,}323{,}744{,}561$ \\
    News (English) $1\%$ & $26{,}244{,}234{,}449$ & $26{,}445{,}407$ & $45{,}117{,}552$ & $5{,}554{,}488$  \\

    \addlinespace[0.5em]
    \rowcolor{gray!20}
    \multicolumn{5}{c}{\textit{Normalized by bilinguialism}} \\
\end{tabular}}
    \caption{Number (in terms of token counts) and proportions of English (\textsc{en}), non-English (\textsc{nen}), bilingual (\textsc{bil}), and translation (\textsc{tra}) instances for each source in PaLM's dataset mixture. Bilingual and translation instances are found within all of PaLM's sources except News articles.}
    \label{tab:per_source_counts}
\end{table*}

%% file: acl2023.bbl
\begin{thebibliography}{33}
\expandafter\ifx\csname natexlab\endcsname\relax\def\natexlab#1{#1}\fi

\bibitem[{Agrawal et~al.(2022)Agrawal, Zhou, Lewis, Zettlemoyer, and
  Ghazvininejad}]{Agrawal2022IncontextES}
Sweta Agrawal, Chunting Zhou, Mike Lewis, Luke Zettlemoyer, and Marjan
  Ghazvininejad. 2022.
\newblock \href {https://doi.org/10.48550/ARXIV.2212.02437} {In-context
  examples selection for machine translation}.

\bibitem[{Blevins and Zettlemoyer(2022)}]{Blevins2022LanguageCH}
Terra Blevins and Luke Zettlemoyer. 2022.
\newblock \href {https://aclanthology.org/2022.emnlp-main.233} {Language
  contamination helps explains the cross-lingual capabilities of {E}nglish
  pretrained models}.
\newblock In \emph{Proceedings of the 2022 Conference on Empirical Methods in
  Natural Language Processing}, pages 3563--3574, Abu Dhabi, United Arab
  Emirates. Association for Computational Linguistics.

\bibitem[{Bojar et~al.(2014)Bojar, Buck, Federmann, Haddow, Koehn, Leveling,
  Monz, Pecina, Post, Saint-Amand, Soricut, Specia, and
  Tamchyna}]{bojar-etal-2014-findings}
Ond{\v{r}}ej Bojar, Christian Buck, Christian Federmann, Barry Haddow, Philipp
  Koehn, Johannes Leveling, Christof Monz, Pavel Pecina, Matt Post, Herve
  Saint-Amand, Radu Soricut, Lucia Specia, and Ale{\v{s}} Tamchyna. 2014.
\newblock \href {https://doi.org/10.3115/v1/W14-3302} {Findings of the 2014
  workshop on statistical machine translation}.
\newblock In \emph{Proceedings of the Ninth Workshop on Statistical Machine
  Translation}, pages 12--58, Baltimore, Maryland, USA. Association for
  Computational Linguistics.

\bibitem[{Botha et~al.(2017)Botha, Pitler, Ma, Bakalov, Salcianu, Weiss,
  McDonald, and Petrov}]{botha-etal-2017-natural}
Jan~A. Botha, Emily Pitler, Ji~Ma, Anton Bakalov, Alex Salcianu, David Weiss,
  Ryan McDonald, and Slav Petrov. 2017.
\newblock \href {https://doi.org/10.18653/v1/D17-1309} {Natural language
  processing with small feed-forward networks}.
\newblock In \emph{Proceedings of the 2017 Conference on Empirical Methods in
  Natural Language Processing}, pages 2879--2885, Copenhagen, Denmark.
  Association for Computational Linguistics.

\bibitem[{Brown et~al.(2020)Brown, Mann, Ryder, Subbiah, Kaplan, Dhariwal,
  Neelakantan, Shy~am, Sastry, Askell, Agarwal, H~erbert Voss, Krueger,
  Henighan, Child, Ramesh, Ziegler, Wu, Winter, Chen, Sigler, Litwin, Gray,
  Chess, Clark, Berner, McCa~ndlish, Radford, Sutskever, and
  Amodei}]{NEURIPS2020_1457c0d6}
Tom Brown, Benjamin Mann, Nick Ryder, Melani~e Subbiah, Jared~D Kaplan,
  Prafulla Dhariwal, Arvind Neelakantan, Pranav Shy~am, Girish Sastry, Amanda
  Askell, Sandhini Agarwal, Ariel H~erbert Voss, Gretchen Krueger, Tom
  Henighan, Rewon Child, Aditya Ramesh, Daniel Ziegler, Jeffrey Wu, Chris
  Winter, Clemens a nd~Hesse, Mark Chen, Eric Sigler, Mateusz Litwin, Scott
  Gray, Benjamin Chess, Jack Clark, Christopher Berner, Sam McCa~ndlish, Alec
  Radford, Ilya Sutskever, and Dario Amodei. 2020.
\newblock \href
  {https://proceedings.neurips.cc/paper/2020/file/1457c0d6bfcb4967418bf
  b8ac142f64a-Paper.pdf} {Language models are few-shot learners}.
\newblock In \emph{Advances in Neural Information Processing Systems},
  volume~33, pages 1877--1901. Curran Associates, Inc.

\bibitem[{Caswell et~al.(2020)Caswell, Breiner, van Esch, and
  Bapna}]{caswell-etal-2020-language}
Isaac Caswell, Theresa Breiner, Daan van Esch, and Ankur Bapna. 2020.
\newblock \href {https://doi.org/10.18653/v1/2020.coling-main.579} {Language
  {ID} in the wild: Unexpected challenges on the path to a thousand-language
  web text corpus}.
\newblock In \emph{Proceedings of the 28th International Conference on
  Computational Linguistics}, pages 6588--6608, Barcelona, Spain (Online).
  International Committee on Computational Linguistics.

\bibitem[{Chowdhery et~al.(2022)Chowdhery, Narang, Devlin, Bosma, Mishra,
  Roberts, Barham, Chung, Sutton, Gehrmann, Schuh, Shi, Tsvyashchenko, Maynez,
  Rao, Barnes, Tay, Shazeer, Prabhakaran, Reif, Du, Hutchinson, Pope, Bradbury,
  Austin, Isard, Gur-Ari, Yin, Duke, Levskaya, Ghemawat, Dev, Michalewski,
  Garcia, Misra, Robinson, Fedus, Zhou, Ippolito, Luan, Lim, Zoph, Spiridonov,
  Sepassi, Dohan, Agrawal, Omernick, Dai, Pillai, Pellat, Lewkowycz, Moreira,
  Child, Polozov, Lee, Zhou, Wang, Saeta, Diaz, Firat, Catasta, Wei,
  Meier-Hellstern, Eck, Dean, Petrov, and Fiedel}]{Chowdhery2022PaLMSL}
Aakanksha Chowdhery, Sharan Narang, Jacob Devlin, Maarten Bosma, Gaurav Mishra,
  Adam Roberts, Paul Barham, Hyung~Won Chung, Charles Sutton, Sebastian
  Gehrmann, Parker Schuh, Kensen Shi, Sasha Tsvyashchenko, Joshua Maynez,
  Abhishek Rao, Parker Barnes, Yi~Tay, Noam Shazeer, Vinodkumar Prabhakaran,
  Emily Reif, Nan Du, Ben Hutchinson, Reiner Pope, James Bradbury, Jacob
  Austin, Michael Isard, Guy Gur-Ari, Pengcheng Yin, Toju Duke, Anselm
  Levskaya, Sanjay Ghemawat, Sunipa Dev, Henryk Michalewski, Xavier Garcia,
  Vedant Misra, Kevin Robinson, Liam Fedus, Denny Zhou, Daphne Ippolito, David
  Luan, Hyeontaek Lim, Barret Zoph, Alexander Spiridonov, Ryan Sepassi, David
  Dohan, Shivani Agrawal, Mark Omernick, Andrew~M. Dai,
  Thanumalayan~Sankaranarayana Pillai, Marie Pellat, Aitor Lewkowycz, Erica
  Moreira, Rewon Child, Oleksandr Polozov, Katherine Lee, Zongwei Zhou, Xuezhi
  Wang, Brennan Saeta, Mark Diaz, Orhan Firat, Michele Catasta, Jason Wei,
  Kathy Meier-Hellstern, Douglas Eck, Jeff Dean, Slav Petrov, and Noah Fiedel.
  2022.
\newblock \href {https://doi.org/10.48550/ARXIV.2204.02311} {Palm: Scaling
  language modeling with pathways}.

\bibitem[{Creswell and Clark(2017)}]{creswell}
John~W. Creswell and Vicki L.~Plano Clark. 2017.
\newblock Designing and conducting mixed methods research.
\newblock \emph{Sage Publications}.

\bibitem[{Devlin et~al.(2019)Devlin, Chang, Lee, and
  Toutanova}]{devlin-etal-2019-bert}
Jacob Devlin, Ming-Wei Chang, Kenton Lee, and Kristina Toutanova. 2019.
\newblock \href {https://doi.org/10.18653/v1/N19-1423} {{BERT}: Pre-training of
  deep bidirectional transformers for language understanding}.
\newblock In \emph{Proceedings of the 2019 Conference of the North {A}merican
  Chapter of the Association for Computational Linguistics: Human Language
  Technologies, Volume 1 (Long and Short Papers)}, pages 4171--4186,
  Minneapolis, Minnesota. Association for Computational Linguistics.

\bibitem[{Dodge et~al.(2021)Dodge, Sap, Marasovi{\'c}, Agnew, Ilharco,
  Groeneveld, Mitchell, and Gardner}]{dodge-etal-2021-documenting}
Jesse Dodge, Maarten Sap, Ana Marasovi{\'c}, William Agnew, Gabriel Ilharco,
  Dirk Groeneveld, Margaret Mitchell, and Matt Gardner. 2021.
\newblock \href {https://doi.org/10.18653/v1/2021.emnlp-main.98} {Documenting
  large webtext corpora: A case study on the colossal clean crawled corpus}.
\newblock In \emph{Proceedings of the 2021 Conference on Empirical Methods in
  Natural Language Processing}, pages 1286--1305, Online and Punta Cana,
  Dominican Republic. Association for Computational Linguistics.

\bibitem[{Feng et~al.(2022)Feng, Yang, Cer, Arivazhagan, and
  Wang}]{feng-etal-2022-language}
Fangxiaoyu Feng, Yinfei Yang, Daniel Cer, Naveen Arivazhagan, and Wei Wang.
  2022.
\newblock \href {https://doi.org/10.18653/v1/2022.acl-long.62}
  {Language-agnostic {BERT} sentence embedding}.
\newblock In \emph{Proceedings of the 60th Annual Meeting of the Association
  for Computational Linguistics (Volume 1: Long Papers)}, pages 878--891,
  Dublin, Ireland. Association for Computational Linguistics.

\bibitem[{Goyal et~al.(2022)Goyal, Gao, Chaudhary, Chen, Wenzek, Ju, Krishnan,
  Ranzato, Guzm{\'a}n, and Fan}]{goyal-etal-2022-flores}
Naman Goyal, Cynthia Gao, Vishrav Chaudhary, Peng-Jen Chen, Guillaume Wenzek,
  Da~Ju, Sanjana Krishnan, Marc{'}Aurelio Ranzato, Francisco Guzm{\'a}n, and
  Angela Fan. 2022.
\newblock \href {https://doi.org/10.1162/tacl_a_00474} {The {F}lores-101
  evaluation benchmark for low-resource and multilingual machine translation}.
\newblock \emph{Transactions of the Association for Computational Linguistics},
  10:522--538.

\bibitem[{Jiang et~al.(2020)Jiang, Xu, Araki, and
  Neubig}]{jiang-etal-2020-know}
Zhengbao Jiang, Frank~F. Xu, Jun Araki, and Graham Neubig. 2020.
\newblock \href {https://doi.org/10.1162/tacl_a_00324} {How can we know what
  language models know?}
\newblock \emph{Transactions of the Association for Computational Linguistics},
  8:423--438.

\bibitem[{Koehn et~al.(2020)Koehn, Chaudhary, El-Kishky, Goyal, Chen, and
  Guzm{\'a}n}]{koehn-etal-2020-findings}
Philipp Koehn, Vishrav Chaudhary, Ahmed El-Kishky, Naman Goyal, Peng-Jen Chen,
  and Francisco Guzm{\'a}n. 2020.
\newblock \href {https://aclanthology.org/2020.wmt-1.78} {Findings of the {WMT}
  2020 shared task on parallel corpus filtering and alignment}.
\newblock In \emph{Proceedings of the Fifth Conference on Machine Translation},
  pages 726--742, Online. Association for Computational Linguistics.

\bibitem[{Kreutzer et~al.(2022)Kreutzer, Caswell, Wang, Wahab, van Esch,
  Ulzii-Orshikh, Tapo, Subramani, Sokolov, Sikasote, Setyawan, Sarin, Samb,
  Sagot, Rivera, Rios, Papadimitriou, Osei, Suarez, Orife, Ogueji, Rubungo,
  Nguyen, M{\"u}ller, M{\"u}ller, Muhammad, Muhammad, Mnyakeni, Mirzakhalov,
  Matangira, Leong, Lawson, Kudugunta, Jernite, Jenny, Firat, Dossou, Dlamini,
  de~Silva, {\c{C}}abuk~Ball{\i}, Biderman, Battisti, Baruwa, Bapna, Baljekar,
  Azime, Awokoya, Ataman, Ahia, Ahia, Agrawal, and
  Adeyemi}]{kreutzer-etal-2022-quality}
Julia Kreutzer, Isaac Caswell, Lisa Wang, Ahsan Wahab, Daan van Esch,
  Nasanbayar Ulzii-Orshikh, Allahsera Tapo, Nishant Subramani, Artem Sokolov,
  Claytone Sikasote, Monang Setyawan, Supheakmungkol Sarin, Sokhar Samb,
  Beno{\^\i}t Sagot, Clara Rivera, Annette Rios, Isabel Papadimitriou, Salomey
  Osei, Pedro~Ortiz Suarez, Iroro Orife, Kelechi Ogueji, Andre~Niyongabo
  Rubungo, Toan~Q. Nguyen, Mathias M{\"u}ller, Andr{\'e} M{\"u}ller,
  Shamsuddeen~Hassan Muhammad, Nanda Muhammad, Ayanda Mnyakeni, Jamshidbek
  Mirzakhalov, Tapiwanashe Matangira, Colin Leong, Nze Lawson, Sneha Kudugunta,
  Yacine Jernite, Mathias Jenny, Orhan Firat, Bonaventure F.~P. Dossou, Sakhile
  Dlamini, Nisansa de~Silva, Sakine {\c{C}}abuk~Ball{\i}, Stella Biderman,
  Alessia Battisti, Ahmed Baruwa, Ankur Bapna, Pallavi Baljekar, Israel~Abebe
  Azime, Ayodele Awokoya, Duygu Ataman, Orevaoghene Ahia, Oghenefego Ahia,
  Sweta Agrawal, and Mofetoluwa Adeyemi. 2022.
\newblock \href {https://doi.org/10.1162/tacl_a_00447} {Quality at a glance: An
  audit of web-crawled multilingual datasets}.
\newblock \emph{Transactions of the Association for Computational Linguistics},
  10:50--72.

\bibitem[{Kudo and Richardson(2018)}]{kudo-richardson-2018-sentencepiece}
Taku Kudo and John Richardson. 2018.
\newblock \href {https://doi.org/10.18653/v1/D18-2012} {{S}entence{P}iece: A
  simple and language independent subword tokenizer and detokenizer for neural
  text processing}.
\newblock In \emph{Proceedings of the 2018 Conference on Empirical Methods in
  Natural Language Processing: System Demonstrations}, pages 66--71, Brussels,
  Belgium. Association for Computational Linguistics.

\bibitem[{Lin et~al.(2021)Lin, Mihaylov, Artetxe, Wang, Chen, Simig, Ott,
  Goyal, Bhosale, Du, Pasunuru, Shleifer, Koura, Chaudhary, O'Horo, Wang,
  Zettlemoyer, Kozareva, Diab, Stoyanov, and Li}]{Lin2021FewshotLW}
Xi~Victoria Lin, Todor Mihaylov, Mikel Artetxe, Tianlu Wang, Shuohui Chen,
  Daniel Simig, Myle Ott, Naman Goyal, Shruti Bhosale, Jingfei Du, Ramakanth
  Pasunuru, Sam Shleifer, Punit~Singh Koura, Vishrav Chaudhary, Brian O'Horo,
  Jeff Wang, Luke Zettlemoyer, Zornitsa Kozareva, Mona Diab, Ves Stoyanov, and
  Xian Li. 2021.
\newblock Few-shot learning with multilingual language models.
\newblock \emph{ArXiv}, abs/2112.10668.

\bibitem[{Lu et~al.(2020)Lu, Ge, Shi, and Zhang}]{lu-etal-2020-alibaba}
Jun Lu, Xin Ge, Yangbin Shi, and Yuqi Zhang. 2020.
\newblock \href {https://aclanthology.org/2020.wmt-1.111} {{A}libaba submission
  to the {WMT}20 parallel corpus filtering task}.
\newblock In \emph{Proceedings of the Fifth Conference on Machine Translation},
  pages 979--984, Online. Association for Computational Linguistics.

\bibitem[{Lu et~al.(2018)Lu, Lv, Shi, and Chen}]{lu-etal-2018-alibaba}
Jun Lu, Xiaoyu Lv, Yangbin Shi, and Boxing Chen. 2018.
\newblock \href {https://doi.org/10.18653/v1/W18-6481} {{A}libaba submission to
  the {WMT}18 parallel corpus filtering task}.
\newblock In \emph{Proceedings of the Third Conference on Machine Translation:
  Shared Task Papers}, pages 917--922, Belgium, Brussels. Association for
  Computational Linguistics.

\bibitem[{Papineni et~al.(2002)Papineni, Roukos, Ward, and
  Zhu}]{papineni-etal-2002-bleu}
Kishore Papineni, Salim Roukos, Todd Ward, and Wei-Jing Zhu. 2002.
\newblock \href {https://doi.org/10.3115/1073083.1073135} {{B}leu: a method for
  automatic evaluation of machine translation}.
\newblock In \emph{Proceedings of the 40th Annual Meeting of the Association
  for Computational Linguistics}, pages 311--318, Philadelphia, Pennsylvania,
  USA. Association for Computational Linguistics.

\bibitem[{Petroni et~al.(2019)Petroni, Rockt{\"a}schel, Riedel, Lewis, Bakhtin,
  Wu, and Miller}]{petroni-etal-2019-language}
Fabio Petroni, Tim Rockt{\"a}schel, Sebastian Riedel, Patrick Lewis, Anton
  Bakhtin, Yuxiang Wu, and Alexander Miller. 2019.
\newblock \href {https://doi.org/10.18653/v1/D19-1250} {Language models as
  knowledge bases?}
\newblock In \emph{Proceedings of the 2019 Conference on Empirical Methods in
  Natural Language Processing and the 9th International Joint Conference on
  Natural Language Processing (EMNLP-IJCNLP)}, pages 2463--2473, Hong Kong,
  China. Association for Computational Linguistics.

\bibitem[{Popovi{\'c}(2015)}]{popovic-2015-chrf}
Maja Popovi{\'c}. 2015.
\newblock \href {https://doi.org/10.18653/v1/W15-3049} {chr{F}: character
  n-gram {F}-score for automatic {MT} evaluation}.
\newblock In \emph{Proceedings of the Tenth Workshop on Statistical Machine
  Translation}, pages 392--395, Lisbon, Portugal. Association for Computational
  Linguistics.

\bibitem[{Post(2018)}]{post-2018-call}
Matt Post. 2018.
\newblock \href {https://doi.org/10.18653/v1/W18-6319} {A call for clarity in
  reporting {BLEU} scores}.
\newblock In \emph{Proceedings of the Third Conference on Machine Translation:
  Research Papers}, pages 186--191, Brussels, Belgium. Association for
  Computational Linguistics.

\bibitem[{Radford et~al.(2018)Radford, Wu, Child, Luan, Amodei, and
  Sutskever}]{Radford2019LanguageMA}
Alec Radford, Jeffrey Wu, Rewon Child, David Luan, Dario Amodei, and Ilya
  Sutskever. 2018.
\newblock \href
  {https://d4mucfpksywv.cloudfront.net/better-language-models/language-models.pdf}
  {Language models are unsupervised multitask learners}.

\bibitem[{Raffel et~al.(2019)Raffel, Shazeer, Roberts, Lee, Narang, Matena,
  Zhou, Li, and Liu}]{Raffel2019ExploringTL}
Colin Raffel, Noam Shazeer, Adam Roberts, Katherine Lee, Sharan Narang, Michael
  Matena, Yanqi Zhou, Wei Li, and Peter~J. Liu. 2019.
\newblock \href {https://doi.org/10.48550/ARXIV.1910.10683} {Exploring the
  limits of transfer learning with a unified text-to-text transformer}.

\bibitem[{Reynolds and McDonell(2021)}]{Reynolds2021PromptPF}
Laria Reynolds and Kyle McDonell. 2021.
\newblock \href {https://doi.org/10.1145/3411763.3451760} {Prompt programming
  for large language models: Beyond the few-shot paradigm}.
\newblock In \emph{Extended Abstracts of the 2021 CHI Conference on Human
  Factors in Computing Systems}, CHI EA '21, New York, NY, USA. Association for
  Computing Machinery.

\bibitem[{Sanh et~al.(2021)Sanh, Webson, Raffel, Bach, Sutawika, Alyafeai,
  Chaffin, Stiegler, Scao, Raja, Dey, Bari, Xu, Thakker, Sharma, Szczechla,
  Kim, Chhablani, Nayak, Datta, Chang, Jiang, Wang, Manica, Shen, Yong, Pandey,
  Bawden, Wang, Neeraj, Rozen, Sharma, Santilli, Fevry, Fries, Teehan, Bers,
  Biderman, Gao, Wolf, and Rush}]{Sanh2021MultitaskPT}
Victor Sanh, Albert Webson, Colin Raffel, Stephen~H. Bach, Lintang Sutawika,
  Zaid Alyafeai, Antoine Chaffin, Arnaud Stiegler, Teven~Le Scao, Arun Raja,
  Manan Dey, M~Saiful Bari, Canwen Xu, Urmish Thakker, Shanya~Sharma Sharma,
  Eliza Szczechla, Taewoon Kim, Gunjan Chhablani, Nihal Nayak, Debajyoti Datta,
  Jonathan Chang, Mike Tian-Jian Jiang, Han Wang, Matteo Manica, Sheng Shen,
  Zheng~Xin Yong, Harshit Pandey, Rachel Bawden, Thomas Wang, Trishala Neeraj,
  Jos Rozen, Abheesht Sharma, Andrea Santilli, Thibault Fevry, Jason~Alan
  Fries, Ryan Teehan, Tali Bers, Stella Biderman, Leo Gao, Thomas Wolf, and
  Alexander~M. Rush. 2021.
\newblock \href {https://doi.org/10.48550/ARXIV.2110.08207} {Multitask prompted
  training enables zero-shot task generalization}.

\bibitem[{Shazeer(2019)}]{shazeer-2019-fast-transformer}
Noam Shazeer. 2019.
\newblock \href {https://doi.org/10.48550/ARXIV.1911.02150} {Fast transformer
  decoding: One write-head is all you need}.

\bibitem[{Shin et~al.(2020)Shin, Razeghi, Logan~IV, Wallace, and
  Singh}]{shin-etal-2020-autoprompt}
Taylor Shin, Yasaman Razeghi, Robert~L. Logan~IV, Eric Wallace, and Sameer
  Singh. 2020.
\newblock \href {https://doi.org/10.18653/v1/2020.emnlp-main.346}
  {{A}uto{P}rompt: {E}liciting {K}nowledge from {L}anguage {M}odels with
  {A}utomatically {G}enerated {P}rompts}.
\newblock In \emph{Proceedings of the 2020 Conference on Empirical Methods in
  Natural Language Processing (EMNLP)}, pages 4222--4235, Online. Association
  for Computational Linguistics.

\bibitem[{Shorten and Smith(2017)}]{shorten}
Allison Shorten and Joanna Smith. 2017.
\newblock Mixed methods research: Expanding the evidence base.
\newblock \emph{Evidence-Based Nursing}.

\bibitem[{Vaswani et~al.(2017)Vaswani, Shazeer, Parmar, Uszkoreit, Jones,
  Gomez, Kaiser, and Polosukhin}]{Vaswani2017Attention}
Ashish Vaswani, Noam Shazeer, Niki Parmar, Jakob Uszkoreit, Llion Jones,
  Aidan~N Gomez, \L~ukasz Kaiser, and Illia Polosukhin. 2017.
\newblock \href
  {https://proceedings.neurips.cc/paper/2017/file/3f5ee243547dee91fbd053c1c4a845aa-Paper.pdf}
  {Attention is all you need}.
\newblock In \emph{Advances in Neural Information Processing Systems},
  volume~30. Curran Associates, Inc.

\bibitem[{Vilar et~al.(2022)Vilar, Freitag, Cherry, Luo, Ratnakar, and
  Foster}]{vilar2022prompting}
David Vilar, Markus Freitag, Colin Cherry, Jiaming Luo, Viresh Ratnakar, and
  George Foster. 2022.
\newblock \href {https://doi.org/10.48550/ARXIV.2211.09102} {Prompting palm for
  translation: Assessing strategies and performance}.

\bibitem[{Zhang et~al.(2018)Zhang, Riesa, Gillick, Bakalov, Baldridge, and
  Weiss}]{zhang-etal-2018-fast-compact}
Yuan Zhang, Jason Riesa, Daniel Gillick, Anton Bakalov, Jason Baldridge, and
  David Weiss. 2018.
\newblock \href {https://doi.org/10.18653/v1/D18-1030} {A fast, compact,
  accurate model for language identification of codemixed text}.
\newblock In \emph{Proceedings of the 2018 Conference on Empirical Methods in
  Natural Language Processing}, pages 328--337, Brussels, Belgium. Association
  for Computational Linguistics.

\end{thebibliography}
